\documentclass[journal,twoside,web]{ieeecolor}
\usepackage{generic}
\usepackage{algorithmic}
\usepackage{graphicx}
\usepackage{algorithm,algorithmic}
\usepackage{hyperref}
\hypersetup{hidelinks=true}
\usepackage{textcomp}
\usepackage{cite}
\usepackage{amsmath,amssymb,amsfonts}
\usepackage{algorithm}
\usepackage{textcomp}
\usepackage{xcolor}
\usepackage{multirow}
\usepackage{graphicx} % 记得加宏包
\usepackage[justification=centering]{caption}
\usepackage{cleveref}

\usepackage{booktabs}
\usepackage{tabularx}
\usepackage{float}
\usepackage{pifont}
\usepackage{epstopdf}

\usepackage{soul}
\usepackage{xcolor}
\usepackage{adjustbox}
\definecolor{highlightyellow}{rgb}{1,1,0.6}
\definecolor{highlightgreen}{rgb}{0.6,1,0.6}

\soulregister{\cite}{1}

\newcommand*{\defeq}{\mathrel{\vcenter{\baselineskip0.5ex \lineskiplimit0pt
			\hbox{\footnotesize.}\hbox{\footnotesize.}}}%
	=}

\def\BibTeX{{\rm B\kern-.05em{\sc i\kern-.025em b}\kern-.08em
    T\kern-.1667em\lower.7ex\hbox{E}\kern-.125emX}}
\begin{document}
\title{FedGSCA: Medical Federated Learning with Global Sample Selector and Client Adaptive Adjuster under Label Noise}
\author{Mengwen Ye, \IEEEmembership{Member, IEEE}, Yingzi Huangfu, Shujian Gao, Wei Ren, Weifan Liu, Zekuan Yu
\thanks{This work was partially supported by the National Natural Science Foundation of China under Grant 82103964. }
\thanks{Mengwen Ye, Yingzi Huangfu, Shujian Gao and Zekuan Yu are with the Academy for Engineering and Technology, Fudan University, Shanghai 200438, China (e-mails: mwye22@m.fudan.edu.cn; fyzhuang23@m.fudan.edu.cn; sjgao24@m.fudan.edu.cn; yzk@fudan.edu.cn).}
\thanks{Wei Ren is a professor in the School of Computer Science, China University of Geosciences, Wuhan 430074, China (e-mail: weirencs@cug.edu.cn).}
\thanks{Weifan Liu is a lecturer in the College of Science, Beijing Forestry University, Beijing 100083, China (e-mail: weifanliu@bjfu.edu.cn).}
\thanks{(Corresponding authors: Zekuan Yu)}}

\maketitle

\begin{abstract}
Federated Learning (FL) emerged as a solution for collaborative medical image classification while preserving data privacy. However, label noise, which arises from inter-institutional data variability, can cause training instability and degrade model performance. Existing FL methods struggle with noise heterogeneity and the imbalance in medical data. Motivated by these challenges, we propose FedGSCA, a novel framework for enhancing robustness in noisy medical FL. FedGSCA introduces a Global Sample Selector that aggregates noise knowledge from all clients, effectively addressing noise heterogeneity and improving global model stability. Furthermore, we develop a Client Adaptive Adjustment (CAA) mechanism that combines adaptive threshold pseudo-label generation and Robust Credal Labeling Loss. CAA dynamically adjusts to class distributions, ensuring the inclusion of minority samples and carefully managing noisy labels by considering multiple plausible labels. This dual approach mitigates the impact of noisy data and prevents overfitting during local training, which improves the generalizability of the model. We evaluate FedGSCA on one real-world colon slides dataset and two synthetic medical datasets under various noise conditions, including symmetric, asymmetric, extreme, and heterogeneous types. The results show that FedGSCA outperforms the state-of-the-art methods, excelling in extreme and heterogeneous noise scenarios. Moreover, FedGSCA demonstrates significant advantages in improving model stability and handling complex noise, making it well-suited for real-world medical federated learning scenarios.
\end{abstract}


\section{Introduction}
\label{sec:introduction}
\IEEEPARstart{F}{ederated} 
Learning (FL) offers a promising solution to the challenges of data privacy and availability in medical image classification by enabling multiple healthcare institutions to collaboratively train a global model without sharing data directly\cite{b1,bc1}. However, most existing FL frameworks assume that client datasets are annotated with clean labels, an assumption that is rarely met in real-world clinical settings\cite{b3}. High-quality annotation requires considerable expertise, and mislabeling is common due to high feature similarity and variations in image quality. As a result, label noise is a frequent and unavoidable issue in medical image datasets\cite{RN706}.

Traditional methods for managing noisy labels, which often rely on centralized data access, are not applicable in FL due to these privacy concerns. And label noise in federated learning is particularly challenging due to heterogeneous annotation quality across clients. Some clients have high-quality annotations, while others deal with noisy labels caused by limited resources\cite{b4}. This variability in noise rates and types among clients can impair the performance of simple aggregation methods like FedAvg\cite{b5}, propagating the negative impact of noisy labels and leading to model overfitting. Despite extensive research on handling label noise in centralized learning settings, the problem of label noise in medical FL remains underexplored. Therefore, there is an urgent need for robust FL frameworks capable of handling label noise to support reliable medical image classification tasks.

Managing noise heterogeneity, which represents the varying noise distributions among clients, poses a significant challenge in medical federated learning with noisy labels (F-LNL)\cite{bc7}. In such scenarios, different clients may have distinct noise characteristics arising from diverse factors, such as variations in data collection processes, annotation quality, or resource constraints. Recent F-LNL approaches filter noisy data locally, but these methods often rely heavily on each client's limited data, which may lack sufficient samples to accurately model local noise distributions and unable to leverage global noise patterns\cite{b8,b9}. This mismatch between local and global updates leads to instability, referring to the convergence fluctuations caused by noise heterogeneity , which in turn affects the reliability and convergence of the model. To address this, we propose the Global Sample Selector (GSS) in our FedGSCA framework. GSS aggregates sample selection information from all clients using a Gaussian Mixture Model (GMM), enabling clients to share noise-handling strategies. This global knowledge helps each hospital more accurately identify noisy samples, improving local sample selectors across clients. As a result, the GSS mitigates the risk of instability caused by noise heterogeneity and reduces the likelihood of overfitting to noisy labels.

Class imbalance is another common challenge in medical diagnostics that can amplify annotation problems. Minority classes—such as rare diseases or early-stage lesions—are pivotal to clinical outcomes, and mislabeling or overlooking them can lead to severe misdiagnoses or missed diagnoses. Furthermore, label noise disproportionately affects these smaller classes, compounding the negative impact of both imbalance and noisy annotations. Traditional methods use fixed thresholds for pseudo-labeling, which can exclude minority class samples, worsening the imbalance\cite{b11, bc11}. To mitigate this problem, our Adaptive Threshold Pseudo-Label (ATP) Generation mechanism dynamically adjusts the threshold for each class based on its distribution, ensuring the full utilization of minority class samples. However, replacing a potentially incorrect label with a pseudo label can be hasty in the early stages of training, as it risks reinforcing the effects of label noise rather than mitigating them\cite{b12}. To address this, we introduce Robust Credal Labeling (RCL) Loss which considers a set of plausible labels (a credal set) for each noisy sample, preventing the model from overcommitting to potentially incorrect labels during the early stages of training. This technique is vital in medical image classification, where subtle differences between conditions can lead to ambiguity in labeling. By considering multiple label possibilities, RCL gradually guides the learning process toward more reliable outcomes. As a part of CAA, RCL Loss ensures more stable learning across clients and contributes to the overall improvement of the global model.

In summary, our contributions are as follows:
\begin{itemize}
\item[$\bullet$] We introduce a novel GSS in medical F-LNL, which aggregates noise information across clients. This mechanism allows the effective sharing of noise-handling knowledge among hospitals, mitigating the impact of noise heterogeneity and enhancing the stability and robustness of the global model.
\item[$\bullet$] To maximize the utilization of local data, mitigate class imbalance, and enhance robustness, we propose the CAA mechanism, which includes (1) Dynamic Data Utilization that generates adaptive pseudo-labels for noisy samples based on class-specific thresholds, and (2) RCL Loss that optimally selects labels from a credal set to reduce the impact of noise.
\item[$\bullet$] We conduct extensive experiments on three widely-used medical image datasets under different noise scenarios, including symmetric, asymmetric, and heterogeneous noise rates and types. The results demonstrate that our FedCAGS framework outperforms the state-of-the-art (SOTA) methods.
\end{itemize}

\section{Related Work}
\subsection{Noisy Label Learning}
In recent years, the challenge of learning from noisy labels has garnered significant attention due to its profound impact on model performance and generalization ability. Traditional machine learning methods have achieved some success in mitigating this issue\cite{b13,b15}. Despite their exceptional representational capabilities, deep learning methods are notably more vulnerable to the detrimental impacts of label noise\cite{b17}. Research efforts in deep learning directed towards tackling noisy labels have predominantly converged into five distinct strategies\cite{b18}: Robust Architecture\cite{b19}, Robust Regularization\cite{b21}, Robust Loss Function\cite{b26}, Loss Adjustment\cite{b28}, Sample Selection\cite{b33}. However, these methods share a common limitation: they primarily focus on handling noise at the local level, such as by adjusting model architecture, loss functions, or sample selection, while overlooking the importance of global noise handling. This is particularly problematic in FL environments, where local noise processing methods often struggle to address the heterogeneous noise distributions across clients, making the global model susceptible to the negative impact of noisy samples from individual clients.

In the medical field,  researchers have proposed several solutions for handling noisy labels, including regularized loss functions and image normalization strategies\cite{b35}, dual uncertainty estimation techniques\cite{b36}, and hybrid modules such as Data Cleaning, Adaptive Negative Learning, and Sharpening-Aware Minimization\cite{b37}, curriculum learning paradigms\cite{b39}. These diverse approaches have been specifically tailored to address various medical image classification tasks, focusing not only on the issue of noisy labels but also on the unique visual characteristics associated with each pathology. However, the increasing demands for data privacy protection, cross-institutional collaboration, and addressing data heterogeneity in the medical field have rendered traditional methods for handling noisy labels inadequate, leading researchers to turn to federated learning.

\subsection{Federated Learning with Noisy Labels}
In classical federated learning algorithms, Federated Averaging (FedAvg) performs a weighted aggregation of client parameters\cite{b5}, while FedProx enhances convergence stability through the introduction of a proximal term and dynamic iterations\cite{b41}. FedRobust addresses distributional shifts caused by device heterogeneity and variations in data distribution\cite{b42}. The traditional F-LNL, such as Client Selection and Data Selection, typically require additional clean data to enhance the selection process and rely on a robust initial model to prevent error accumulation\cite{b43}. Recent advances in F-LNL primarily have focused on the integration of global and local information\cite{b44}, the introduction of adaptive mechanisms\cite{b46}, the identification and modeling of noisy data\cite{b47}, and the optimization of noise filtering and weighting strategies\cite{b49}, \cite{b50}, \cite{b51}, \cite{b52}. For example, Federated Learning with Label Noise Correction (FedCorr) balances local model adaptation to local data with global model consistency through adaptive local regularization and global model fine-tuning\cite{b44}.  However, most of these methods primarily focus on improving local noise handling and heavily rely on the division between clean and noisy clients.

Research on F-LNL  in medical imaging classification tasks remains limited. The Federated Graph Purification (FedGP) framework purifies noisy graphs at the client level and expands them using topological knowledge to generate pseudo-labels, while global centroid aggregation at the server level creates a collaboratively optimized classifier\cite{b54}. The Adaptive Sample Weighting Federated Learning method, incorporates joint training into the FL, improving diagnostic accuracy for brain MRI analysis\cite{b55}. 

Our research aims to develop a simple but efficient architecture that mitigates the negative effects of noisy labels on model performance without the need for explicit identification of noisy datasets. To validate the effectiveness of our approach, we conduct extensive experiments under both homogeneous and heterogeneous noise conditions.

\section{Method}
\begin{figure*}[t!]
    \centering
    \includegraphics[width=0.75\linewidth]{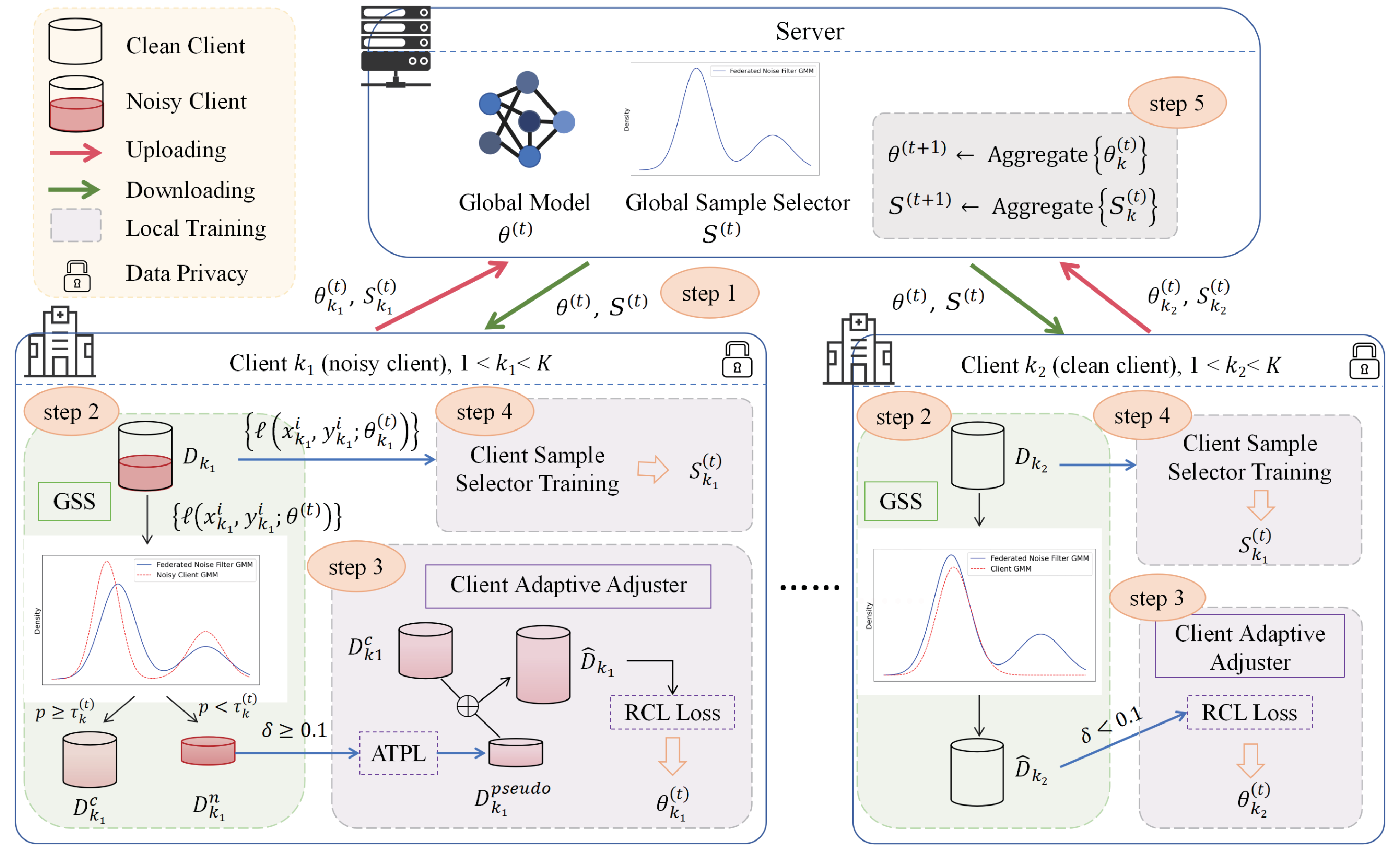}
    \caption{Overview of the FedGSCA framework. Client $k_1$ represents a noisy client, while Client $k_2$ represents a clean client. Each client uses a GMM-based approach to separate clean and noisy samples. Client $k_1$ with high noise proportions applies ATPL before RCL-based training, while $k_2$ with low noise is trained directly with the RCL Loss. The server aggregates both the local model, and updates the global model parameters and the sample selector parameters. GSS, Global Sample Selector; ATPL, Adaptive Threshold Pseudo-Label Generation; RCL, Robust Credal Labelling. }
    \label{fig1:Figure1}
\end{figure*}
In this section, we introduce FedCAGS, which comprises two key components: the GSS and the CAA. The GSS enhances the model's adaptability to heterogeneous noise. The CAA enhances model robustness through two key components: Dynamic Data Utilization and the RCL Loss.

We assume a medical federated learning scenario, involving a central server and \(K\) local clients denoted by \(S\). For any client \(k \in S\), its private dataset is denoted as \(D_k = \{(x_i, y_i)\}_{i=1}^{n_k}\), where \(n_k\) represents the number of samples in the \(k\)-th client's dataset, and \(y_i \in \mathcal{Y}\) is the label for sample \(x_i\), applicable to a classification task with \(c\) classes. Clients may contain clean or noisy labels, without assuming a fixed noise distribution.
The description of symbols in the manuscript as shown
in Table \ref{table-symbol}.
\begin{table}
    \centering
    \caption{The Description of Symbols}
    \label{table-symbol}
    \setlength{\tabcolsep}{8pt}
    \renewcommand{\arraystretch}{1.2} 
    \begin{tabular}{|p{30pt}|p{165pt}|}
        \hline
        \pmb{Symbol} & \pmb{Description} \\
        \hline
        \( S \)        & A medical federated learning scenario. \\
        \( K \)        & The number of local clients. \\
        \( M \)        & The set of all clients \( k \in M \). \\
        \( D_k \)      & Any local client’s private dataset \(\{(x_i, y_i)\}_{i=1}^{n_k}\). \\
        \( D_k^c \)    & The clean subset of \( D_k \). \\
        \( D_k^n \)    & The noisy subset of \( D_k \). \\
        \( D_k^{\text{pseudo}} \) & A new subset of the pseudo-labeled samples. \\
        \( \hat{D}_k \) & The updated training dataset of the \( k \)-th client. \\
        \( n_k \)      & The number of samples in the \( k \)-th client’s dataset. \\
        \( \delta_k \) & The noise level of the \( k \)-th client. \\
        \( \theta^{(t)} \) & The global model parameters. \\
        \( \theta_k^{(t)} \) & The local model parameters of the \( k \)-th client. \\
        \( S_k^{(t)} \) & The parameters of the \( k \)-th CSS in the \( t \)-th communication round \((\mu_k^t, \sigma_k^t, \pi_k^t)\). \\
        \( S^{(t+1)} \) & The Global Sample Selector parameters aggregated from all clients’ CSS parameters. \\
        \( \tau_k^{(t)} \) & Selector coefficient. \\
        \( \text{Avg}_c \) & The average confidence \(\text{Avg}_c\) for each class \( c \) using the clean samples. \\
        \( \zeta_0 \)  & The initial threshold for pseudo-label generation. \\
        \( \zeta_c \)  & The adaptive threshold \(\zeta_c\) for each class \( c \). \\
        \( \Pi_i(y) \) & The possibility distribution for a sample \( x_i \) and label \( y_i \). \\
        \( Q_{\Pi_i} \) & The credal set constructed based on the possibility distribution \( \Pi_i(y) \). \\
        \( \alpha \)   & The label relaxation parameter. \\
        \( \beta \)    & The confidence threshold. \\
        \( T \)        & The total number of iterations rounds used for training. \\
        \hline
    \end{tabular}
    \label{tab1}
\end{table}

\subsection{Overview of FedGSCA Framework}
An overview of the FedGSCA framework is shown in Fig. \ref{alg:FedGSCA}, and the general process is shown in Alg. 1.
%为了实现优化目标，FedGSCA第t个通信回合的训练过程包括以下6个步骤：
% step 1: 每一个客户端 \(k \in S\) download the parameters of global model thetat and GSS S（t）from the server.
% step 2: 对于每一个\(k \in S\) ，根据S（t）将Dk分为干净子集Dkclean和噪声子集Dknoisy
%step 3: 对于每一个客户端。training process 包括更新全局模型 with the Dkclean和根据rcl更新客户端模型
%step 3: 对干净子集D_k^{c}进进行Mixup增强，结合RCL损失函数来训练本地模型theta，得到更新的local model parameters。
%step 4：通过更新的本地模型theta计算D_k的每个样本的损失，根据每个样本的损失值训练一个更新后的client sample selector
%step 5：the server 通过 FedAvg aggregates所有客户端的本地模型参数以获得更新全局模型参数，类似的，服务器聚合所有客户端的client sample selector以获得更新的GSS
The FedGSCA training process for the $t$-th communication round consists of the following five steps:

\pmb{Step 1}: Each client $k \in \mathcal{M}$ downloads the parameters of the global model $\theta^{(t)}$ and the GSS $S^{(t)}$ from the server.

\pmb{Step 2} (GSS): For each client $k \in \mathcal{M}$, the local dataset $D_k$ is divided into a clean subset $D_k^{c}$ and a noisy subset $D_k^{n}$ based on $S^{(t)}$ and selector coefficient $\tau_k^{(t)}$.

\pmb{Step 3} (CAA): Compute $\delta_k$ for client $k$. If $\delta_k \geq 0.1$, train the $k$-th local model using the RCL loss on the union of $D_k^c$ and the pseudo-labeled subset $D_k^{\text{pseudo}}$. If $\delta_k < 0.1$, train the model using $D_k$. This yields the updated local model parameters $\theta_k^{(t)}$.

\pmb{Step 4} (GSS): Calculate the loss using the updated local model $\theta_k^{(t)}$. Based on the loss values of each sample, train an updated client sample selector (CSS) $S_k^{(t)}$.

\pmb{Step 5}: The server aggregates $\{\theta_k^{(t)} \mid k \in \mathcal{M}\}$ using FedAvg to update global model $\theta^{(t+1)}$. Similarly, the server aggregates $\{S_k^{(t)} \mid k \in \mathcal{M}\}$ to update the GSS $S^{(t+1)}$.

%The details of \pmb{Step 2}, \pmb{Step 4}, and \pmb{Step 5} are provided in section B. Global Sample Selector, while the detailed description of \pmb{Step 3} is given in section C. Dynamic Data Utilization and D. Robust Credal Labelling Loss.

\subsection{Global Sample Selector}
%GSS 它对所有客户端中清洁和有噪音样本的分布进行全局建模。具体来说，我们首先进行client sample selector训练，以获得局部估计的GMM参数，学习本地噪声分布。然后在服务器端聚合这些本地的GMM参数，构建GSS。最后再把GSS广播给各个客户端，根据全局GMM参数选择出干净样本。
GSS is designed to model the distribution of clean and noisy samples across all clients. It starts with training each client's CSS $S_k^{(t)}$ using a GMM to estimate local noise distribution. The server aggregates the CSS parameters of all clients to get the GSS $S^{(t+1)}$. Finally, $S^{(t+1)}$ is broadcasted to all clients for selecting clean samples.
\subsubsection{Client Sample Selector Training}
%模型倾向于优先学习干净样本before fitting label noise，干净样本的损失值通常较小。因此，对于对于第 k 个客户端，在第 t 轮训练中，可以通过拟合每个样本的损失分布来构建一个本地两成分的高斯混合模型（GMM），以模拟本地的干净样本和噪声样本分布。损失值的计算公式为：，在Client Sample Selector Training中，损失函数值被认为服从一个两成分的高斯混合模型（GMM），其中一个成分对应于干净样本，另一个成分对应于噪声样本。每个客户端都会根据其私有数据集中样本的损失值来训练一个本地 GMM 模型。
%客户端的数据可能混合了干净和带噪声的样本。为了保证模型训练的准确性，我们可以基于噪声样本和干净样本的损失值差异，通过高斯混合模型区分噪声样本与干净样本。
%首先我们计算客户端$k$的每个样本的损失值。通常，带有噪声标签的样本会有较大的损失值，而干净标签的样本损失值则相对较小。因此，可以利用样本的损失值来区分干净样本和带噪声的样本。为了实现这个目标，客户端为其本地数据训练一个高斯混合模型，用来建模不同样本的损失值分布。然后，我们需要通过计算后验概率来判断每个样本是干净的概率还是带噪声的概率。为了估计 GMM 的参数 ，我们在每个客户端使用期望最大化（EM）算法。注意我们用来自服务器的全局参数来初始化本地的EM算法参数来加速收敛。 EM算法由两步组成。通过反复进行 E 步和 M 步，模型可以逐步收敛到最佳的参数估计，最终得到最能表示本地数据噪声分布的 GMM 模型。最后客户端会将训练好的本地过滤器参数上传至服务器。
Each client calculates the sample loss using cross-entropy based on $\theta_k^{(t)}$, where noisy samples tend to show higher loss\cite{b56}. Based on this, a GMM is trained on local data $D_k$ to distinguish clean \( g = 1 \) and noisy samples \( g = 2 \)\cite{d1,d2}. Assuming that the loss values follow a mixture of two Gaussian distributions, the conditional probability is defined as follows:
\begin{equation}
P(\ell(x, y; \theta^{(t)}_k) \mid z = g) = N(\ell(x, y; \theta^{(t)}_k); \mu^{(t)}_{kg}, \sigma^{(t)}_{kg})
\label{eq2}
\end{equation}
where \( \mu^{(t)}_{kg} \) and \( \sigma^{(t)}_{kg} \) denote the mean and variance of the \( g \)-th component's loss distribution, respectively. The parameters of the $k$-th CSS in the $t$-th communication round are denoted by:
\begin{equation}
S^{(t)}_k = (\mu^{(t)}_k, \sigma^{(t)}_k, \pi^{(t)}_k),
\label{eq:gmm_parameters}
\end{equation}
where \( \mu^t_k = (\mu^t_{k1}, \mu^t_{k2}) \) and \( \sigma^t_k = (\sigma^t_{k1}, \sigma^t_{k2}) \) are the mean and variance vectors for the clean and noisy components, and \( \pi^t_k = (\pi^t_{k1}, \pi^t_{k2}) \) represents the prior probabilities for each component, with \( \pi^t_{k1} + \pi^t_{k2} = 1 \). The $k$-th client computes the posterior probability for each sample to determine the likelihood of it being clean or noisy. The posterior probability is calculated as follows:
\begin{equation}
\gamma_{kg}(x, y; \theta_k^{(t)}) = \frac{P(\ell(x, y; \theta_k^{(t)})|z = g) P(z = g)}{\sum_{g' = 1}^{2} P(\ell(x, y; \theta_k^{(t)})|z = g') P(z = g')}
\label{eq3}
\end{equation}

In order to estimate the parameters of the GMM, we employ the Expectation-Maximization (EM) algorithm. It is important to note that we initialize the CSS $S_k^{(t)}$ using the parameter of GSS $S^{(t)}$ to accelerate convergence. By iteratively performing the E and M steps, the model converges to the optimal parameter estimates, ultimately yielding a GMM that best represents the noise distribution in the local data. Finally, the client uploads 3 numerical matrices, $S_k^{(t)} = (\mu^t_k, \sigma^t_k, \pi^t_k)$, to the server for protecting the data privacy.

\subsubsection{GSS Aggregation and Clean Samples Selection}
%服务器接收所有客户端的样本选择器参数并进行全局聚合，以便在下一轮更新全局噪声过滤器，如下所示。
After each communication round, the server computes a weighted average of each CSS parameters $\{S_k^{(t)} \mid k \in \mathcal{M}\}$ to update the GSS parameters \( S^{(t+1)} = (\mu^{(t+1)}, \sigma^{(t+1)}, \pi^{(t+1)}) \), as follows:
\begin{equation}
\begin{aligned}
    \mu^{(t+1)} &= \sum_{k \in \mathcal{M}} \frac{n_k}{\sum_{k \in \mathcal{M}} n_k} \mu_{k}, \\
    \sigma^{(t+1)} &= \sum_{k \in \mathcal{M}} \frac{n_k}{\sum_{k \in \mathcal{M}} n_k} \sigma_{k}, \\
    \pi^{(t+1)} &= \sum_{k \in \mathcal{M}} \frac{n_k}{\sum_{k \in \mathcal{M}} n_k} \pi_{k}.
\end{aligned}
\label{eq5}
\end{equation}
At the start of the ($t$+1)-th communication round, the $k$-th client downloads the updated GSS \( S^{(t+1)} \) and updated global model \( \theta^{(t+1)} \) for clean sample selection\cite{b9},\cite{d3}.

Then we compute the selector coefficient $\tau_k^{(t)}$ based on the loss distribution statistics of client k:

\begin{equation}
\tau_k^{(t)} = \min\left(0.5 \cdot \left(1 + \frac{\sigma_k^{L(t)}}{\mu_k^{L(t)} + \epsilon}\right), 0.8\right)
\label{eq:adaptive_threshold}
\end{equation}
where $\mu_k^{L(t)}$ and $\sigma_k^{L(t)}$ represent the mean and standard deviation of loss values for client $k$ at round $t$, the choice of 0.8 as an upper bound is based on empirical observations to prevent overly strict clean sample selection, and $\epsilon$ is set to $1 \times 10^{-6}$. 
 We then estimate the probability that a sample $x$ from $D_k$ is clean through its posterior probability for the “clean” component, as shown below: 
\begin{equation}
p(\text{“clean”} \mid x, y; \theta^{(t)}) = P(z = 1 \mid x, y; \theta^{(t)})
\label{eq6}
\end{equation}

Samples exceeding $\tau_k^{(t)}$ are labeled as "clean" and added to the clean subset $D_k^c$: 
\begin{equation}
D_k^c = \{(x, y) \mid p(\text{“clean”} \mid x, y; \theta^{(t)}) \geq \tau_k^{(t)}, \forall (x, y) \in D_k \}
\label{eqnew}
\end{equation}
If the probability is less than $\tau_k^{(t)}$, the sample is assigned to the noisy subset $D_k^n$.Eq.\eqref{eq:adaptive_threshold} leverages the empirical observation that clients with more noisy samples typically exhibit more dispersed loss distributions compared to clients with fewer noisy samples\cite{b56}. When the loss distribution is highly dispersed, $\tau_k^{(t)}$ will be larger, making it more stringent to classify samples as clean. Conversely, when the loss distribution is less dispersed, $\tau_k^{(t)}$ will be smaller, making it easier to classify samples as clean.

\subsection{Dynamic Data Utilization}
%为了在本地训练时避免模型记忆标签噪声，我们引入了自适应阈值和双置信度策略，该方法充分利用了昂贵的医学图像数据并缓解医学图像领域中的类别不平衡问题。
\subsubsection{Noise Level Computation}
The noise level of the $k$-th client is computed as:
%\begin{equation}
%\delta_k = \frac{|D_n^k|}{|D_k|}
%\label{eq7}
%\end{equation}
\begin{equation}
\delta_k = \frac{|D_k^n|}{|D_k|}
\label{eq7}
\end{equation}
where \( |D_k^n| \) represents the number of noisy samples, and \( |D_k| \) is the total dataset size for client \( k \). If \( \delta_k > 0.1 \), we discard the labels of noisy samples from \( |D_k^n| \) to prevent the local model from learning incorrect labels.
\subsubsection{Adaptive Threshold Pseudo-Label Generation}
%目前的方法大部分均用固定阈值来生成伪标签，但由于固定阈值较高，导致大量样本无法参与模型训练，这对非常珍贵和本就不多的医学图像数据集非常不友好。尤其是在医学图像数据不平衡的情况下，模型往往更偏向多数类，从而进一步降低对少数类的置信度。为了解决这些问题我们计算xxx
Fixed thresholds often exclude valuable samples in medical imaging datasets due to imbalance. To address this, we use an adaptive threshold mechanism, ensuring more samples are included in training while mitigating class imbalance. To begin, we firstly compute the average confidence \( \text{Avg}_c \) for each class \( c \) using the clean samples in \( D_k^c \) :
\begin{equation}
\text{Avg}_c = \frac{1}{|D_k^c|} \sum_{x_i \in D_k^c} \mathbb{I}(\arg\max(p(x_i; \theta^{(t)})) = c) \cdot m_i
\end{equation}
where \( m_i = \max(p(x_i; \theta^{(t)})) \), and \( p(x_i; \theta^{(t)}) \) represents the global model’s predicted probability distribution for sample \( x_i \).
The adaptive threshold \( \zeta_c \) for each class \( c \) is calculated as:
\begin{equation}
\zeta_c = \zeta_0 \times \frac{\text{Avg}_c}{\max(\text{Avg}_c)}
\end{equation}
where \( \zeta_0 \) is the initial threshold. By lowering the threshold for classes with lower average confidence, the method ensures that pseudo-labels are still generated for minority classes, thus mitigating the effects of class imbalance in medical datasets. If the global model’s maximum predicted probability \( \max(p(x; \theta^{(t)})) \) of noisy sample \( x \) exceeds the class-specific threshold, a pseudo-label is assigned:
\begin{equation}
\hat{y} = \arg\max p(x; \theta^{(t)})
\label{eq10}
\end{equation}

\subsubsection{Client Training Dataset}
The pseudo-labeled samples form a new subset \( D_k^{\text{pseudo}} \), defined as:
\begin{equation}
D_k^{\text{pseudo}} = \{(x, \hat{y}) \, | \, x \in D_k^n, \, \max(p(x; \theta^{(t)})) \geq \zeta_{\hat{y}} \}
\label{eq10}
\end{equation}
For further optimizing the local model, the updated training dataset for client $k$ is consequently constructed to fully utilize noisy samples as follows:
\begin{equation}
\hat{D}_k =
\begin{cases}
D_k^c \cup D_k^{\text{pseudo}}, & \text{if } \delta_k \geq 0.1, \\
D_k, & \text{if } \delta_k < 0.1.
\end{cases}
\end{equation}

\subsection{Robust Credal Labelling Loss}
After constructing the updated dataset $\hat{D}_k$ for the $k$-th client, we apply the RCL Loss to optimize the $k$-th local model for robustness against label noise\cite{lienen2023conformal}.

In traditional probabilistic learning settings, deterministic target labels $y_i \in \mathcal{Y}$ are transformed into degenerate probability distributions $p_{y_i} \in \mathbb{P}(\mathcal{Y})$, where all probability mass is assigned to the observed label $y_i$, i.e. $p_{y_i}(y_i) = 1$ and $p_{y_i}(y) = 0$ for all $y \neq y_i$. This approach works well in clean data environments but can lead to poor generalization in noisy environments, forcing the model to memorize incorrect labels. To address this issue, we introduce credal sets to better manage label uncertainty, particularly on noisy datasets\cite{ lienen2024mitigating}, \cite{ wang2024creinnscredalsetintervalneural}. Credal set $Q_{\pi_i}$ represents a set of plausible probability distributions over the label space $\mathcal{Y}$, providing the model with the flexibility to hedge its predictions between the observed label and alternative candidates.

\subsubsection{Credal Set and Probability Projection}
Based on possibility theory\cite{b58}, we define the possibility distribution $\pi_i(y)$ as follows: 
\begin{equation} 
    \pi_i(y) = \begin{cases} 
        1 & \text{if } y = y_i \text{ or } \hat{p}(y \mid x_i; \theta_k^{(t)}) \geq \beta \\
        \alpha & \text{otherwise}
    \end{cases}\label{eqpi}
\end{equation}
where \( \beta, \alpha \in [0,1] \). The label relaxation parameter \( \alpha \) allows the local model to relax its confidence for noisy samples, meaning that even for low-probability labels, their possibility is not entirely ruled out. Drawing on our previous work\cite{arxiv}, \(\beta\) is updated using cosine decay.

Based on \( \pi_i \), the credal set \( Q_{\pi_i} \) is defined as the set of all probability distributions that satisfy the constraints imposed by \( \pi_i \)\cite{caprio2024credallearningtheory}, defined as:
\begin{equation} 
Q_{\pi_i} \defeq \Big\{ p \in \mathbb{P}(\mathcal{Y}) \, \vert \, \forall \, Y \subseteq \mathcal{Y} : \, \sum_{y \in Y} p(y) \leq \max_{y \in Y} \pi_i(y) \Big\}\label{eqQPII}
\end{equation}

To handle uncertainty during training, we introduce the projected probability distribution \( p_r \), derived by projecting the local model \(\theta_k^{(t)}\)'s predicted distribution \( \hat{p}(y \mid x_i; \theta_k^{(t)}) \) onto the boundary of the credal set \( Q_{\pi_i} \). The projected probability distribution \( p_r \) is defined as follows:
\begin{equation}
    p^r(y) = \begin{cases}
(1 - \alpha) \cdot \frac{ \hat{p}(y \mid x_i; \theta_k^{(t)})}{\sum_{y' \in \mathcal{Y} : \pi_i(y')=1} \hat{p}(y' \mid x_i; \theta_k^{(t)})} & \text{if } \pi_i(y) = 1 \\
\alpha \cdot \frac{ \hat{p}(y \mid x_i; \theta_k^{(t)})}{\sum_{y' \in \mathcal{Y} : \pi_i(y')=\alpha} \hat{p}(y' \mid x_i; \theta_k^{(t)})} & \text{otherwise}
\end{cases}\label{eq}
\end{equation}
This projection adjusts the model's predictions based on the credal set, ensuring that the model incorporates the uncertainty expressed in the possibility distribution.
\begin{algorithm}[t]
\caption{FedGSCA}
\label{alg:FedGSCA}
\pmb{Input:} $\{D_k \mid k \in \mathcal{M}\}$; $\theta$; $S$; $T$; $\beta, \alpha \in [0,1]$\\
\pmb{Output:} $\theta$; $S$\\
\pmb{Server executes:}
\begin{algorithmic}[1]
    \STATE Initialize $\theta^{(0)}$ and $S^{(0)}$;
    \FOR{each round $t \in \{1, 2, \dots, T\}$}
        \FOR{each client $k \in \{1, 2, \dots, K\}$ \pmb{in parallel}}
            \STATE $\theta_k, S_k \gets$ \pmb{LocalUpdate}($t, k, \theta, S$);
        \ENDFOR
        \STATE Update $\theta$ with $\{\theta_k\mid k \in \mathcal{M}\}$ via FedAvg;
        \STATE Update $S$ with $\{S_k\mid k \in \mathcal{M}\}$ using (\ref{eq5});
        \ENDFOR
    \RETURN well-trained $\theta$ and $S$.
\end{algorithmic}
\pmb{LocalUpdate($t, k, \theta, S$):}
\begin{algorithmic}[1]
    \STATE Calculate $\tau_k^{(t)}$ using \eqref{eq:adaptive_threshold};
    \STATE Divide dataset $D_k$ into $D_k^c$ and $D_k^n$ using \eqref{eqnew};
    \STATE $\delta_k \gets \frac{|D_k^n|}{|D_k|}$;
    \IF{$\delta_k \geq 0.1$}
        \STATE Generate pseudo-labels $D_k^{\text{pseudo}}$ for $D_k^n$ with adaptive threshold $\zeta_c$ using (\ref{eq10});
        \STATE $\hat{D}_k \gets D_k^c \cup D_k^{\text{pseudo}}$;
    \ELSE
        \STATE $\hat{D}_k \gets D_k$
    \ENDIF
    \FOR{local epoch $e = 1, 2, \dots, E$}
    \STATE Calculate $\pi$ as specified in \eqref{eqpi}
    \STATE Calculate the credal set $Q_\pi$ in \eqref{eqQPII}
    \STATE Optimize $\theta_k$ using \eqref{eqLOSS}
    \ENDFOR

    \WHILE{$S_k$ is not converged}
        \STATE Calculate $\gamma_{kg}(x, y; \theta_k^{(t)})$ in (3)
        \STATE \(\mu_{kg} = \frac{\sum_{(x,y) \in \mathcal{D}_k} \gamma_{kg}(x,y;\theta_k) \cdot \ell(x,y;\theta_k)}{\sum_{(x,y) \in \mathcal{D}_k} \gamma_{kg}(x,y;\theta_k)}\)
        \STATE \(\sigma_{kg} = \frac{\sum_{(x,y) \in \mathcal{D}_k} \gamma_{kg}(x,y;\theta_k) \cdot \left[ \ell(x,y;\theta_k) - \mu_{kg} \right]^2}{\sum_{(x,y) \in \mathcal{D}_k} \gamma_{kg}(x,y;\theta_k)}\)
        \STATE \(\pi_{kg} = \frac{\sum_{(x,y) \in \mathcal{D}_k} \gamma_{kg}(x,y;\theta_k)}{n_k}\)
    \ENDWHILE
    \RETURN updated $\theta_k$, $S_k$;
\end{algorithmic}
\end{algorithm}
\subsubsection{Loss Function Optimization}
The loss function selects the probability distribution \( p \in Q_{\pi_i} \) that minimizes the loss function \( \mathcal{L}(p, \hat{p}) \), taking into account the entire credal set for the samples in \( \hat{D}_k \). The loss is defined as:
\begin{equation}
    \mathcal{L}^*(Q_{\pi} , \hat{p}) = \begin{cases}
        0  & \text{if } \hat{p} \in Q_{\pi} \\
        D_{KL}(p^r \, || \, \hat{p}) & \text{otherwise} 
    \end{cases} \, ,
    \label{eqLOSS}
\end{equation}
where the Kullback-Leibler Divergence \( D_{KL} \) measures the divergence between the projected probability \( p^r \) and the predicted probability \( \hat{p} \).
\section{EXPERIMENTS}
\subsection{Experimental Setup}
\subsubsection{Datasets}
To evaluate medical FL label noise, we conduct experiments on three widely used medical imaging datasets:
\begin{itemize}
\item[$\bullet$] Kvasir-Capsule dataset\cite{b59}: For a fair comparison, we followed the FedGP procedure for the Kvasir-Capsule dataset. We split the endoscopic dataset into training, validation, and test sets in a 7:1:2 ratio.

\item[$\bullet$] OIA-ODIR dataset\cite{b60}: This dataset contains fundus images of 3500 patients for training and 1500 patients for testing. For multi-label images, we only use the first label.

\item[$\bullet$] Real-world chaoyang dataset\cite{b61}: The dataset consists of 1816 normal, 1163 serrated, 2244 adenocarcinoma, and 937 adenoma colon slides. Of these, 705 normal, 321 serrated, 840 adenocarcinoma, and 273 adenoma colon slides with consensus results from all three pathologists are used for testing. For approximately 40$\%$ of the training samples where three doctors provided inconsistent annotations, we randomly selected one pathologist’s opinion and ensured that each label’s class distribution matched FedGP’s approach. 
\end{itemize}

The training set of Kvasir-Capsule and OIA-ODIR datasets are equally divided among four clients for the FL scenario. For the Chaoyang dataset, we divided the training set evenly into three clients, following FedGP.
\subsubsection{Label Noise Settings}
To simulate real-world medical label noise, we applied the Clinical Knowledge-Based Asymmetric (CK-Asymm.) Noise proposed in our previous work\cite{arxiv}, which introduces noise based on clinical misdiagnosis patterns. In addition, we incorporated symmetric and pairflip label noise to ensure a fair comparison with other methods.  Table \ref{table1-1:kvasir_misdiagnosis_candidates} and Table \ref{table1-2:oia_odir_misdiagnosis_candidates} provide details on the misdiagnosis set \(S(y_i)\) based on domain knowledge.
\begin{table}[h]
    \centering
    \caption{Misdiagnosis Candidates for Kvasir-Capsule Dataset}
    \label{table1-1:kvasir_misdiagnosis_candidates}
    \begin{tabular}{>{\centering\arraybackslash}p{70pt}>{\centering\arraybackslash}p{140pt}}
        \toprule
        \pmb{Class} & \pmb{Misdiagnosis Set \(S(y_i)\)} \\
        \midrule
        Normal                 & Random choice  \\
        Ileocecal Valve        & Mucosal Atrophy, Pylorus  \\
        Mucosal Atrophy        & Ileocecal Valve, Erosion  \\
        Pylorus                & Ulcer, Foreign Body       \\
        Angiectasia            & Erosion, Blood            \\
        Ulcer                  & Pylorus, Foreign Body     \\
        Foreign Body           & Ulcer, Lymphangiectasia   \\
        Lymphangiectasia       & Erosion, Foreign Body     \\
        Erosion                & Mucosal Atrophy, Angiectasia \\
        Blood                  & Angiectasia, Erosion      \\
        \bottomrule
    \end{tabular}
\end{table}
\begin{table}[h]
    \centering
    \caption{Misdiagnosis Candidates for OIA-ODIR Dataset}
    \label{table1-2:oia_odir_misdiagnosis_candidates}
    \begin{tabular}{>{\centering\arraybackslash}p{80pt}>{\centering\arraybackslash}p{140pt}}
        \toprule
        \pmb{Class} & \pmb{Misdiagnosis Set \(S(y_i)\)} \\
        \midrule
        Normal                 & Random choice \\
        Hypertensive Retinopathy & Glaucoma, Hypertension Complications \\
        Glaucoma               & Cataract, Hypertensive Retinopathy \\
        Cataract               & AMD, Glaucoma  \\
        AMD                    & Cataract, Hypertensive Retinopathy \\
        Hypertension Complications & Pathologic Myopia, AMD \\
        Pathologic Myopia      & Hypertension Complications, Other Diseases/Abnormalities \\
        Other Diseases/Abnormalities & Pathologic Myopia, Hypertension Complications \\
        \bottomrule
    \end{tabular}
\end{table}
\subsubsection{Evaluation metrics}
To assess performance under medical FL label noise, we use F1 score, recall, and precision for a comprehensive comparison of FL methods. Recall measures prediction completeness, precision evaluates quality, and F1 provides a balanced summary. For multi-class diagnosis, metrics are calculated per class and then averaged using macro-average to give equal weight to each category. Higher values in these metrics indicate better diagnostic performance on the three medical datasets.
\subsubsection{Implementation details}
We implemented FedCAGS using PyTorch with the LVM-Med Pretrained ResNet-50 model as the backbone\cite{bc14}. All images were preprocessed to a size of 224 $\times$ 224 and the local training is augmented with random flip and rotation of input images to avoid overfitting. In the comparison of both datasets, the networks in the proposed FedGSCA framework and the SOTA methods are optimized with SGD for $T$ = 100 iterations with a batch size of 128 and a weight decay of $1 \times 10^{-6}$. The initial learning rate is to as $1 \times 10^{-2}$, and the learning rate is divided by 10 at 70$\%$ and 90$\%$ progress of local training to ensure convergence. For adaptive threshold pseudo-label (ATP) generation, the initial threshold \( \zeta_0 \) is 0.8. For robust RCL, we set $\beta_0 = 0.75$, $\beta_1 = 0.55$ and $\alpha = 0.05$. All experiments were conducted on 2 $\times$ NVIDIA GeForce RTX 3090 GPU.

\subsection{Comparison With State-of-the-Art Methods}
We compare the performance of our method with the SOTA methods (i.e., FedAvg\cite{b5}, FedProx\cite{b41}, FedRobust\cite{b42}, FedCorr\cite{b44}, FedGP\cite{b54}, FedFixer\cite{c1}), and FedNoRo\cite{b62}). All methods used the same data splits, client simulations, random seed, and model initialization for a fair comparison.
\subsubsection{Results on the Kvasir-Capsule Dataset}
On the Kvasir-Capsule dataset under different label noise settings, including symmetric, pairflip, and CK-Asymm. noise, as well as heterogeneous noise rates and types, we evaluate the robustness and adaptability of our proposed method across these challenging scenarios.

Under extreme symmetric label noise, as shown in Table \ref{table2:Comparison on Kvasir-Capsule dataset under symmetric label noise}, our method outperforms other approaches. When half of the samples have noisy labels, our approach still achieves the best F1 score (77.39) and the best recall (77.78). For pairflip and CK-Asymm. label noise, the results in Table \ref{table3:Comparison on Kvasir-Capsule dataset under Pairflip and CK-Asymm.} further illustrate the superiority of our method.

\begin{table*}[h]
    \centering
    \caption{Comparison on Kvasir-Capsule dataset under extreme symmetric label noise. Best and second-best results are highlighted and underlined.}
    \label{table2:Comparison on Kvasir-Capsule dataset under symmetric label noise}
    \small  % 整体缩小字号
    \setlength{\tabcolsep}{18pt}  % 调整列间距
    \renewcommand{\arraystretch}{0.8}  % 调整行高
    \begin{tabular}{@{}l*{6}{c}@{}}  % 使用紧凑列格式
        \toprule
        \multirow{2}{*}{\centering Method} & \multicolumn{3}{c}{40\%} & \multicolumn{3}{c}{50\%} \\
        \cmidrule(r){2-4} \cmidrule(l){5-7}
        & F1 & Recall & Prec. & F1 & Recall & Prec. \\
        \midrule
        FedAvg    & 65.37{\scriptsize$\pm$0.43} & 66.84{\scriptsize$\pm$0.54} & 70.02{\scriptsize$\pm$0.60} & 56.81{\scriptsize$\pm$0.38} & 57.12{\scriptsize$\pm$0.50} & 59.39{\scriptsize$\pm$1.03} \\
        FedProx   & 66.79{\scriptsize$\pm$0.29} & 68.21{\scriptsize$\pm$0.56} & 70.35{\scriptsize$\pm$0.73} & 57.51{\scriptsize$\pm$0.45} & 59.29{\scriptsize$\pm$0.63} & 60.00{\scriptsize$\pm$1.06} \\
        FedRobust & 62.59{\scriptsize$\pm$0.39} & 63.24{\scriptsize$\pm$0.51} & 66.67{\scriptsize$\pm$0.44} & 56.05{\scriptsize$\pm$0.39} & 57.21{\scriptsize$\pm$0.54} & 61.13{\scriptsize$\pm$1.12} \\
        FedCorr   & 62.79{\scriptsize$\pm$0.28} & 63.09{\scriptsize$\pm$0.37} & 63.85{\scriptsize$\pm$0.64} & 58.02{\scriptsize$\pm$0.29} & 57.88{\scriptsize$\pm$0.37} & 61.20{\scriptsize$\pm$0.46} \\
        FedGP     & \underline{81.52{\scriptsize$\pm$0.25}} & 80.94{\scriptsize$\pm$0.31} & 83.11{\scriptsize$\pm$0.47} & 76.39{\scriptsize$\pm$0.20} & 77.11{\scriptsize$\pm$0.43} & 77.68{\scriptsize$\pm$0.51} \\
        FedFixer  & 81.21{\scriptsize$\pm$0.30} & \underline{81.12{\scriptsize$\pm$0.32}} & \underline{83.17{\scriptsize$\pm$0.32}} & \underline{77.12{\scriptsize$\pm$0.35}} & \underline{77.52{\scriptsize$\pm$0.51}} & 78.15{\scriptsize$\pm$0.62} \\
        FedNoRo     & 79.48{\scriptsize$\pm$0.28} & 79.73{\scriptsize$\pm$0.36} & 80.66{\scriptsize$\pm$0.48} & 76.17{\scriptsize$\pm$0.27} & 77.30{\scriptsize$\pm$0.49} & \underline{78.18{\scriptsize$\pm$0.67}} \\
        \pmb{Ours} & \pmb{82.47{\scriptsize$\pm$0.21}} & \pmb{81.75{\scriptsize$\pm$0.19}} & \pmb{84.40{\scriptsize$\pm$0.26}} & \pmb{77.39{\scriptsize$\pm$0.23}} & \pmb{77.78{\scriptsize$\pm$0.24}} & \pmb{78.22{\scriptsize$\pm$0.28}} \\
        \bottomrule
    \end{tabular}
\end{table*}

\begin{table*}[h]
   \centering
   \caption{Comparison on Kvasir-Capsule dataset under Pairflip and CK-Asymm. label noise. Best and second best results are highlighted and underlined.}
   \label{table3:Comparison on Kvasir-Capsule dataset under Pairflip and CK-Asymm.}
   \resizebox{\linewidth}{!}{
   \large
   \begin{tabular}{lcccccccccccc}
       \toprule
       \multirow{3}{*}{\centering\vspace{-2mm} Method}& \multicolumn{6}{c}{Pairflip noise} & \multicolumn{6}{c}{CK-Asymm. noise} \\
       \cmidrule(r){2-7} \cmidrule(l){8-13}
       & \multicolumn{3}{c}{10\%} & \multicolumn{3}{c}{20\%} & \multicolumn{3}{c}{10\%} & \multicolumn{3}{c}{20\%} \\
       \cmidrule(r){2-4} \cmidrule(lr){5-7} \cmidrule(lr){8-10}\cmidrule(l){11-13}
       & F1 & Recall & Prec. & F1 & Recall & Prec. & F1 & Recall & Prec. & F1 & Recall & Prec. \\
       \midrule
       FedAvg   & 81.57{\footnotesize$\pm$0.15} & 82.51{\footnotesize$\pm$0.19} & 89.46{\footnotesize$\pm$0.25} & 74.35{\footnotesize$\pm$0.28} & 75.00{\footnotesize$\pm$0.30} & 83.11{\footnotesize$\pm$0.32} & 79.16{\footnotesize$\pm$0.28} & 79.74{\footnotesize$\pm$0.56} & 86.92{\footnotesize$\pm$0.71} & 71.57{\footnotesize$\pm$0.33} & 72.03{\footnotesize$\pm$0.47} & 77.79{\footnotesize$\pm$0.66} \\
       FedProx  & 82.03{\footnotesize$\pm$0.10} & 82.74{\footnotesize$\pm$0.17} & 89.92{\footnotesize$\pm$0.22} & 75.17{\footnotesize$\pm$0.11} & 75.58{\footnotesize$\pm$0.19} & 83.18{\footnotesize$\pm$0.33} & 79.87{\footnotesize$\pm$0.21} & 80.33{\footnotesize$\pm$0.42} & 87.83{\footnotesize$\pm$0.57} & 72.12{\footnotesize$\pm$0.35} & 72.48{\footnotesize$\pm$0.51} & 79.20{\footnotesize$\pm$0.68} \\
       FedRobust & 81.24{\footnotesize$\pm$0.27} & 81.46{\footnotesize$\pm$0.39} & 88.92{\footnotesize$\pm$0.38} & 71.91{\footnotesize$\pm$0.22} & 72.20{\footnotesize$\pm$0.39} & 81.94{\footnotesize$\pm$0.44} & 80.01{\footnotesize$\pm$0.35} & 80.49{\footnotesize$\pm$0.48} & 87.68{\footnotesize$\pm$0.56} & 72.42{\footnotesize$\pm$0.40} & 72.63{\footnotesize$\pm$0.56} & 79.58{\footnotesize$\pm$0.73} \\
       FedCorr  & 82.34{\footnotesize$\pm$0.11} & 83.39{\footnotesize$\pm$0.15} & 90.61{\footnotesize$\pm$0.25} & 74.14{\footnotesize$\pm$0.17} & 74.08{\footnotesize$\pm$0.19} & 77.00{\footnotesize$\pm$0.35} & 79.85{\footnotesize$\pm$0.20} & 78.89{\footnotesize$\pm$0.28} & 87.28{\footnotesize$\pm$0.36} & 71.43{\footnotesize$\pm$0.25} & 71.00{\footnotesize$\pm$0.31} & 74.26{\footnotesize$\pm$0.40} \\
       FedGP    & \underline{88.83{\footnotesize$\pm$0.14}} & \underline{87.85{\footnotesize$\pm$0.14}} & 90.29{\footnotesize$\pm$0.19} & \underline{85.46{\footnotesize$\pm$0.18}} & \underline{85.02{\footnotesize$\pm$0.25}} & 87.74{\footnotesize$\pm$0.31} & \underline{87.68{\footnotesize$\pm$0.11}} & \underline{86.23{\footnotesize$\pm$0.22}} & \underline{88.59{\footnotesize$\pm$0.28}} & \underline{83.57{\footnotesize$\pm$0.14}} & \underline{83.19{\footnotesize$\pm$0.36}} & 85.77{\footnotesize$\pm$0.38} \\
       FedFixer & 83.20{\footnotesize$\pm$0.08} & 83.38{\footnotesize$\pm$0.14} & 87.79{\footnotesize$\pm$0.30} & 82.17{\footnotesize$\pm$0.13} & 81.85{\footnotesize$\pm$0.21} & \underline{88.18{\footnotesize$\pm$0.29}} & 82.14{\footnotesize$\pm$0.26} & 82.10{\footnotesize$\pm$0.27} & 87.32{\footnotesize$\pm$0.27} & 82.77{\footnotesize$\pm$0.31} & 82.24{\footnotesize$\pm$0.32} & \pmb{85.86{\footnotesize$\pm$0.45}} \\
       FedNoRo& 86.64{\footnotesize$\pm$0.25}& 85.80{\footnotesize$\pm$0.28}& \underline{91.17{\footnotesize$\pm$0.29}}& 83.24{\footnotesize$\pm$0.19}& 83.19{\footnotesize$\pm$0.32}& 85.31{\footnotesize$\pm$0.46}& 84.09{\footnotesize$\pm$0.25}& 85.20{\footnotesize$\pm$0.31}& 86.76{\footnotesize$\pm$0.39}& 82.01{\footnotesize$\pm$0.27}& 81.95{\footnotesize$\pm$0.43}& 83.98{\footnotesize$\pm$0.51}
       \\
       \pmb{Ours}     & \pmb{89.40{\footnotesize$\pm$0.06}} & \pmb{88.87{\footnotesize$\pm$0.11}} & \pmb{91.33{\footnotesize$\pm$0.15}} & \pmb{86.16{\footnotesize$\pm$0.07}} & \pmb{85.50{\footnotesize$\pm$0.06}} & \pmb{88.28{\footnotesize$\pm$0.19}} & \pmb{88.22{\footnotesize$\pm$0.08}} & \pmb{88.10{\footnotesize$\pm$0.11}} & \pmb{90.59{\footnotesize$\pm$0.19}} & \pmb{84.10{\footnotesize$\pm$0.09}} & \pmb{83.82{\footnotesize$\pm$0.18}} & \underline{85.77{\footnotesize$\pm$0.17}} \\
       \bottomrule
   \end{tabular}}
\end{table*}
For the heterogeneous noise rate experiments (Table \ref{table4:Comparison on Kvasir-Capsule dataset under label noise with heterogeneous rates.}), the configurations 0-10$\%$-10$\%$-20$\%$ and 0-20$\%$-20$\%$-40$\%$ represent varying noise rates among the four clients. Our method achieves the highest F1 score (88.91) and precision (90.52) under the most challenging setting (0-20$\%$-20$\%$-40$\%$), outperforming FedGP by 2.46$\%$ in F1. Under pairflip noise, our method also leads, demonstrating superior resilience to varying noise rates. 

\begin{table*}[h]
    \centering
    \caption{Comparison on Kvasir-Capsule dataset under label noise with heterogeneous rates. Best and second best results are highlighted and underlined.}
    \label{table4:Comparison on Kvasir-Capsule dataset under label noise with heterogeneous rates.}
    %\resizebox{\linewidth}{!}{
    \resizebox{\linewidth}{!}{
    \large
    \begin{tabular}{lcccccccccccc}
    %\begin{tabular*}{\linewidth}{@{\extracolsep{\fill}}ccccccccccccc@{}}
    %\begin{tabular*}{\linewidth}{@{\extracolsep{\fill}}p{1cm}p{1cm}p{1cm}p{1cm}p{1cm}p{1cm}p{1cm}p{1cm}p{1cm}p{1cm}p{1cm}p{1cm}p{1cm}@{}}
    %\begin{tabular*}{\linewidth}{@{\extracolsep{\fill}}ccccccccccccc@{}}
        \toprule
        \multirow{3}{*}{\centering\vspace{-2mm} Method}& \multicolumn{6}{c}{Symm. noise} & \multicolumn{6}{c}{Pairflip noise
} \\
        \cmidrule(r){2-7} \cmidrule(l){8-13}
        & \multicolumn{3}{c}{0-10\%-10\%-20\%} & \multicolumn{3}{c}{0-20\%-20\%-40\%} & \multicolumn{3}{c}{0-10\%-10\%-20\%} & \multicolumn{3}{c}{0-20\%-20\%-40\%
} \\
        \cmidrule(r){2-4} \cmidrule(lr){5-7} \cmidrule(lr){8-10}\cmidrule(l){11-13}
        & F1& Recall& Prec.& F1& Recall& Prec.& F1& Recall& Prec.& F1& Recall& Prec.
\\
        \midrule
        FedAvg& 85.54{\footnotesize$\pm$0.20}& 84.71{\footnotesize$\pm$0.31}& 90.30{\footnotesize$\pm$0.46}& 80.36{\footnotesize$\pm$0.28}& 80.85{\footnotesize$\pm$0.33}& 85.23{\footnotesize$\pm$0.59}& 82.12{\footnotesize$\pm$0.21}& 81.63{\footnotesize$\pm$0.24}& 80.74{\footnotesize$\pm$0.55}& 77.78{\footnotesize$\pm$0.27}& 78.29{\footnotesize$\pm$0.38}& 77.87{\footnotesize$\pm$0.73}
\\
        FedProx& 85.46{\footnotesize$\pm$0.15}& 84.81{\footnotesize$\pm$0.27}& 91.08{\footnotesize$\pm$0.26}& 81.24{\footnotesize$\pm$0.19}& 81.19{\footnotesize$\pm$0.32}& 85.31{\footnotesize$\pm$0.32}& 82.06{\footnotesize$\pm$0.25}& 81.67{\footnotesize$\pm$0.34}& 81.06{\footnotesize$\pm$0.39}& 77.07{\footnotesize$\pm$0.37}& 77.89{\footnotesize$\pm$0.58}& 77.85{\footnotesize$\pm$0.81}
\\
        FedRobust& 87.53{\footnotesize$\pm$0.22}& 86.17{\footnotesize$\pm$0.30}& 89.34{\footnotesize$\pm$0.41}& 80.49{\footnotesize$\pm$0.31}& 80.33{\footnotesize$\pm$0.50}& 83.83{\footnotesize$\pm$0.79}& 81.15{\footnotesize$\pm$0.27}& 79.50{\footnotesize$\pm$0.48}& 87.52{\footnotesize$\pm$0.52}& 72.64{\footnotesize$\pm$0.36}& 73.11{\footnotesize$\pm$0.52}& 81.25{\footnotesize$\pm$0.84}
\\
        FedCorr& 89.19{\footnotesize$\pm$0.16}& 87.30{\footnotesize$\pm$0.21}& 91.47{\footnotesize$\pm$0.32}& 85.01{\footnotesize$\pm$0.33}& 84.64{\footnotesize$\pm$0.52}& \underline{89.73{\footnotesize$\pm$0.61}}& 84.28{\footnotesize$\pm$0.25}& 83.89{\footnotesize$\pm$0.44}& 91.59{\footnotesize$\pm$0.46}& 76.26{\footnotesize$\pm$0.31}& 76.48{\footnotesize$\pm$0.37}& 84.34{\footnotesize$\pm$0.49}
\\
        FedGP& 88.62{\footnotesize$\pm$0.17}& \underline{88.73{\footnotesize$\pm$0.29}}& 90.92{\footnotesize$\pm$0.35}& \underline{86.45{\footnotesize$\pm$0.31}}& \underline{85.58{\footnotesize$\pm$0.42}}& 88.22{\footnotesize$\pm$0.52}& 86.55{\footnotesize$\pm$0.20}& 85.38{\footnotesize$\pm$0.41}& 88.16{\footnotesize$\pm$0.46}& \underline{83.63{\footnotesize$\pm$0.24}}& 82.60{\footnotesize$\pm$0.45}& 85.39{\footnotesize$\pm$0.67}
\\
        FedFixer& 86.45{\footnotesize$\pm$0.18}& 86.39{\footnotesize$\pm$0.33}& 90.30{\footnotesize$\pm$0.46}& 84.14{\footnotesize$\pm$0.19}& 84.43{\footnotesize$\pm$0.35}& 87.88{\footnotesize$\pm$0.62}& 84.11{\footnotesize$\pm$0.21}& 84.83{\footnotesize$\pm$0.37}& 88.57{\footnotesize$\pm$0.53}& 82.58{\footnotesize$\pm$0.29}& \underline{82.63{\footnotesize$\pm$0.42}}& 88.02{\footnotesize$\pm$0.60}
\\
        FedNoRo& \underline{89.26{\footnotesize$\pm$0.21}}& 88.65{\footnotesize$\pm$0.36}& \underline{91.58{\footnotesize$\pm$0.49}}& 86.28{\footnotesize$\pm$0.27}& 84.76{\footnotesize$\pm$0.38}& 89.69{\footnotesize$\pm$0.57}& \underline{87.09{\footnotesize$\pm$0.26}}& \underline{86.25{\footnotesize$\pm$0.33}}& \underline{91.61{\footnotesize$\pm$0.58}}& 83.12{\footnotesize$\pm$0.29}& 82.62{\footnotesize$\pm$0.44}& \underline{88.08{\footnotesize$\pm$0.68}}
\\
        \pmb{Ours}& \pmb{90.77{\footnotesize$\pm$0.09}}& \pmb{89.73{\footnotesize$\pm$0.17}}& \pmb{92.85{\footnotesize$\pm$0.27}}& \pmb{88.91{\footnotesize$\pm$0.16}}& \pmb{88.03{\footnotesize$\pm$0.19}}& \pmb{90.52{\footnotesize$\pm$0.28}}& \pmb{90.24{\footnotesize$\pm$0.14}}& \pmb{89.70{\footnotesize$\pm$0.28}}& \pmb{91.75{\footnotesize$\pm$0.30}}& \pmb{86.36{\footnotesize$\pm$0.18}}& \pmb{85.25{\footnotesize$\pm$0.28}}& \pmb{88.13{\footnotesize$\pm$0.33}}\\
        \bottomrule
    \end{tabular}}
\end{table*}
\begin{table*}[h]
    \centering
    \caption{Comparison on Kvasir-Capsule dataset under label noise with heterogeneous types. Best and second-best results are highlighted and underlined.}
    \label{table5:Comparison on Kvasir-Capsule dataset under label noise with heterogeneous types}
    \small  % 整体缩小字号
    \setlength{\tabcolsep}{5pt}  % 调整列间距
    \renewcommand{\arraystretch}{0.8}  % 调整行高
    \begin{tabular}{@{}l*{9}{c}@{}}  % 使用紧凑列格式
        \toprule
        \multirow{2}{*}{\centering Method} & \multicolumn{3}{c}{0-S10\%-P10\%-C10\%} & \multicolumn{3}{c}{S10\%-S20\%-P10\%-P20\%} & \multicolumn{3}{c}{0-S40\%-P20\%-C20\%} \\
        \cmidrule(r){2-4} \cmidrule(lr){5-7} \cmidrule(l){8-10}
        & F1 & Recall & Prec. & F1 & Recall & Prec. & F1 & Recall & Prec. \\
        \midrule
        FedAvg    & 82.09{\scriptsize$\pm$0.25} & 82.55{\scriptsize$\pm$0.52} & 90.96{\scriptsize$\pm$0.69} & 80.45{\scriptsize$\pm$0.29} & 81.36{\scriptsize$\pm$0.48} & 79.65{\scriptsize$\pm$0.73} & 75.16{\scriptsize$\pm$0.54} & 75.83{\scriptsize$\pm$0.81} & 75.90{\scriptsize$\pm$1.15} \\
        FedProx   & 82.11{\scriptsize$\pm$0.28}     & 82.92{\scriptsize$\pm$0.49}     & 90.04{\scriptsize$\pm$0.67}     & 80.22{\scriptsize$\pm$0.41}     & 80.89{\scriptsize$\pm$0.70}     & 79.63{\scriptsize$\pm$0.81}     & 75.09{\scriptsize$\pm$0.43}     & 75.78{\scriptsize$\pm$0.71}     & 76.02{\scriptsize$\pm$1.29} \\
        FedRobust & 84.28{\scriptsize$\pm$0.18}     & 83.61{\scriptsize$\pm$0.59}     & 89.15{\scriptsize$\pm$0.83}     & 79.99{\scriptsize$\pm$0.46}     & 80.34{\scriptsize$\pm$0.53}     & 86.67{\scriptsize$\pm$0.97}     & 70.25{\scriptsize$\pm$0.45}     & 71.61{\scriptsize$\pm$0.74}     & 78.57{\scriptsize$\pm$0.31} \\
        FedCorr   & 84.05{\scriptsize$\pm$0.19}     & 85.14{\scriptsize$\pm$0.31}     & 87.76{\scriptsize$\pm$0.49}     & 83.18{\scriptsize$\pm$0.22}     & 83.33{\scriptsize$\pm$0.42}     & 83.23{\scriptsize$\pm$0.53}     & 75.33{\scriptsize$\pm$0.27}     & 75.43{\scriptsize$\pm$0.48}     & 83.26{\scriptsize$\pm$0.65} \\
        FedGP     & \underline{89.10{\scriptsize$\pm$0.20}} & 87.33{\scriptsize$\pm$0.33} & 90.24{\scriptsize$\pm$0.38} & \underline{87.57{\scriptsize$\pm$0.28}} & \underline{86.62{\scriptsize$\pm$0.46}} & 88.46{\scriptsize$\pm$0.49} & \underline{81.20{\scriptsize$\pm$0.31}} & 80.22{\scriptsize$\pm$0.72} & 83.64{\scriptsize$\pm$0.86} \\
        FedFixer  & 83.30{\scriptsize$\pm$0.30}     & 83.27{\scriptsize$\pm$0.37}     & 88.49{\scriptsize$\pm$0.43}     & 82.38{\scriptsize$\pm$0.28}     & 82.25{\scriptsize$\pm$0.44}     & 88.06{\scriptsize$\pm$0.52}     & 80.32{\scriptsize$\pm$0.26}     & 79.62{\scriptsize$\pm$0.54}     & \underline{85.15{\scriptsize$\pm$0.66}} \\
        FedNoRo  & 88.72{\scriptsize$\pm$0.33}     & \underline{87.62{\scriptsize$\pm$0.42}}     & \underline{90.38{\scriptsize$\pm$0.56}}     & 86.68{\scriptsize$\pm$0.37}     & 85.98{\scriptsize$\pm$0.62}     & \underline{89.14{\scriptsize$\pm$0.74}}     & 80.94{\scriptsize$\pm$0.39}     & \underline{81.57{\scriptsize$\pm$0.68}}     & 85.03{\scriptsize$\pm$0.97} \\
        \pmb{Ours} & \pmb{90.03{\scriptsize$\pm$0.11}} & \pmb{89.21{\scriptsize$\pm$0.20}} & \pmb{92.18{\scriptsize$\pm$0.23}} & \pmb{89.07{\scriptsize$\pm$0.14}} & \pmb{88.67{\scriptsize$\pm$0.19}} & \pmb{91.38{\scriptsize$\pm$0.22}} & \pmb{85.71{\scriptsize$\pm$0.19}} & \pmb{84.87{\scriptsize$\pm$0.24}} & \pmb{86.09{\scriptsize$\pm$0.41}} \\
        \bottomrule
    \end{tabular}
\end{table*}

In the heterogeneous label type experiments (Table \ref{table5:Comparison on Kvasir-Capsule dataset under label noise with heterogeneous types}), each client is subject to a different noise type—symmetric (S), pairflip (P), or CK-Asymm. (C). Our method outperforms other SOTA methods, achieving an F1 score of 85.71 and precision of 86.09 in the most challenging scenario (0-S40$\%$-P20$\%$-C20$\%$), surpassing FedGP by 4.51$\%$ in F1, further confirming its robustness in handling mixed noise patterns across clients.
\subsubsection{Results on the OIA-ODIR Dataset}
On the OIA-ODIR dataset, our method shows superior performance under symmetric and pairflip noise conditions with different heterogeneous rates (Table \ref{table6:Comparison on OIA-ODIR dataset under label noise with heterogeneous rates}). For the 0-10$\%$-10$\%$-20$\%$ configuration, our method achieves the highest F1 score (87.30) under symmetric noise and 86.97 under pairflip noise. Even in the most challenging noise configuration (0-20$\%$-20$\%$-40$\%$), our method maintains the best performance with an F1 score of 85.16 for symmetric noise and 83.54 for pairflip noise.
\begin{table*}[h]
    \centering
    \caption{Comparison on OIA-ODIR dataset under label noise with heterogeneous rates. Best and second-best results are highlighted and underlined.} 
    \label{table6:Comparison on OIA-ODIR dataset under label noise with heterogeneous rates}
    \resizebox{\linewidth}{!}{
    \large
    \begin{tabular}{lcccccccccccc}
    %\tiny
    %\begin{tabular*}{\linewidth}{@{\extracolsep{\fill}}p{1cm}p{1cm}p{1cm}p{1cm}p{1cm}p{1cm}p{1cm}p{1cm}p{1cm}p{1cm}p{1cm}p{1cm}p{1cm}@{}}
    %\begin{tabular*}{\linewidth}{@{\extracolsep{\fill}}ccccccccccccc@{}}
        \toprule
        Method& \multicolumn{6}{c}{Symm. noise} & \multicolumn{6}{c}{Pairflip noise
} \\
        \cmidrule(r){2-7} \cmidrule(l){8-13}
        & \multicolumn{3}{c}{0-10\%-10\%-20\%} & \multicolumn{3}{c}{0-20\%-20\%-40\%} & \multicolumn{3}{c}{0-10\%-10\%-20\%} & \multicolumn{3}{c}{0-20\%-20\%-40\%
} \\
        \cmidrule(r){2-4} \cmidrule(lr){5-7} \cmidrule(l){8-10}\cmidrule(l){11-13}
        & F1& Recall& Prec.& F1& Recall& Prec.& F1& Recall& Prec.& F1& Recall& Prec.
\\
        \midrule
        FedAvg& 80.81{\footnotesize$\pm$0.24}& 80.26{\footnotesize$\pm$0.37}& 85.88{\footnotesize$\pm$0.40}& 77.34{\footnotesize$\pm$0.29}& 76.26{\footnotesize$\pm$0.35}& 80.32{\footnotesize$\pm$0.46}& 78.95{\footnotesize$\pm$0.28}& 78.12{\footnotesize$\pm$0.41}& 78.69{\footnotesize$\pm$0.52}& 74.13{\footnotesize$\pm$0.35}& 74.28{\footnotesize$\pm$0.57}& 76.87{\footnotesize$\pm$0.74}
\\
        FedProx& 80.32{\footnotesize$\pm$0.37}& 80.30{\footnotesize$\pm$0.46}& 87.14{\footnotesize$\pm$0.77}& 79.27{\footnotesize$\pm$0.37}& 78.01{\footnotesize$\pm$0.52}& 82.19{\footnotesize$\pm$0.78}& 79.32{\footnotesize$\pm$0.40}& 78.52{\footnotesize$\pm$0.53}& 80.03{\footnotesize$\pm$0.69}& 74.14{\footnotesize$\pm$0.43}& 73.86{\footnotesize$\pm$0.64}& 76.95{\footnotesize$\pm$1.02}
\\
        FedRobust& 82.23{\footnotesize$\pm$0.31}& 82.06{\footnotesize$\pm$0.38}& 87.11{\footnotesize$\pm$0.62}& 79.50{\footnotesize$\pm$0.39}& 79.34{\footnotesize$\pm$0.53}& 84.69{\footnotesize$\pm$0.74}& 78.88{\footnotesize$\pm$0.41}& 78.21{\footnotesize$\pm$0.46}& 83.65{\footnotesize$\pm$0.61}& 70.57{\footnotesize$\pm$0.48}& 70.16{\footnotesize$\pm$0.53}& 79.49{\footnotesize$\pm$0.75}
\\
        FedCorr& 85.34{\footnotesize$\pm$0.33}& 84.03{\footnotesize$\pm$0.48}&  88.02{\footnotesize$\pm$0.52}& 82.17{\footnotesize$\pm$0.36}& 81.60{\footnotesize$\pm$0.44}& 85.08{\footnotesize$\pm$0.57}& 81.61{\footnotesize$\pm$0.29}& 80.59{\footnotesize$\pm$0.40}&  87.04{\footnotesize$\pm$0.42}& 73.97{\footnotesize$\pm$0.37}& 74.45{\footnotesize$\pm$0.46}& 80.94{\footnotesize$\pm$0.58}
\\
        FedGP&  \underline{86.37{\footnotesize$\pm$0.25}}& 85.45{\footnotesize$\pm$0.35}& 87.93{\footnotesize$\pm$0.41}&  83.56{\footnotesize$\pm$0.28}&  \underline{82.41{\footnotesize$\pm$0.36}}& 85.26{\footnotesize$\pm$0.59}&  \underline{84.82{\footnotesize$\pm$0.27}}&  \underline{84.13{\footnotesize$\pm$0.48}}& 86.21{\footnotesize$\pm$0.55}& 80.28{\footnotesize$\pm$0.36}& 80.13{\footnotesize$\pm$0.51}& 84.46{\footnotesize$\pm$0.62}
\\
        FedFixer& 83.17{\footnotesize$\pm$0.29}& 83.62{\footnotesize$\pm$0.44}& 87.18{\footnotesize$\pm$0.57}& 81.03{\footnotesize$\pm$0.31}& 81.29{\footnotesize$\pm$0.38}&  \underline{85.43{\footnotesize$\pm$0.49}}& 81.37{\footnotesize$\pm$0.37}& 81.25{\footnotesize$\pm$0.63}& 84.66{\footnotesize$\pm$0.78}&  80.58{\footnotesize$\pm$0.35}&  80.54{\footnotesize$\pm$0.81}&  \underline{84.80{\footnotesize$\pm$0.90}}
\\
        FedNoRo& 86.08{\footnotesize$\pm$0.35}& \underline{85.49{\footnotesize$\pm$0.47}}& \underline{88.09{\footnotesize$\pm$0.58}}& \underline{83.65{\footnotesize$\pm$0.36}}& 82.32{\footnotesize$\pm$0.49}&  85.37{\footnotesize$\pm$0.62}& 83.99{\footnotesize$\pm$0.31}& 83.78{\footnotesize$\pm$0.49}& \underline{87.87{\footnotesize$\pm$0.52}}&  \underline{81.39{\footnotesize$\pm$0.43}}&  \underline{82.00{\footnotesize$\pm$0.52}}&  84.78{\footnotesize$\pm$0.57}
\\
        \pmb{Ours}& \pmb{87.30{\footnotesize$\pm$0.12}}& \pmb{86.85{\footnotesize$\pm$0.16}}& \pmb{88.38{\footnotesize$\pm$0.18}}& \pmb{85.16{\footnotesize$\pm$0.19}}& \pmb{84.64{\footnotesize$\pm$0.21}}& \pmb{86.01{\footnotesize$\pm$0.23}}& \pmb{86.97{\footnotesize$\pm$0.24}}& \pmb{86.09{\footnotesize$\pm$0.25}}& \pmb{87.95{\footnotesize$\pm$0.27}}& \pmb{83.54{\footnotesize$\pm$0.24}}& \pmb{84.70{\footnotesize$\pm$0.28}}& \pmb{85.34{\footnotesize$\pm$0.29}}\\
        \bottomrule
    \end{tabular}}
\end{table*}
For the heterogeneous noise type experiments (Table \ref{table7:Comparison on OIA-ODIR dataset under label noise with heterogeneous types}), our method also achieves the best results across different noise types. For example, in the most challenging noise configuration (0-S40$\%$-P20$\%$-C20$\%$), our method attains the best F1 score of 82.08, outperforming FedGP (80.15).
\begin{table*}[h]
    \centering
    \caption{Comparison on OIA-ODIR dataset under label noise with heterogeneous types. Best and second-best results are highlighted and underlined.}
    \label{table7:Comparison on OIA-ODIR dataset under label noise with heterogeneous types}
    \small  % 整体缩小字号
    \setlength{\tabcolsep}{5pt}  % 调整列间距
    \renewcommand{\arraystretch}{0.8}  % 调整行高
    \begin{tabular}{@{}l*{9}{c}@{}}  % 使用紧凑列格式
        \toprule
        Method& \multicolumn{3}{c}{0-S10\%-P10\%-C10\%}&\multicolumn{3}{c}{S10\%-S20\%-P10\%-P20\%}& \multicolumn{3}{c}{0-S40\%-P20\%-C20\%}\\
        \cmidrule(r){2-4} \cmidrule(lr){5-7} \cmidrule(l){8-10}
        & F1&Recall& Prec.& F1&Recall& Prec.& F1&Recall& Prec.\\
        \midrule
        FedAvg& 79.31{\scriptsize$\pm$0.31}& 79.40{\scriptsize$\pm$0.36}& 86.06{\scriptsize$\pm$0.49}& 77.63{\scriptsize$\pm$0.33}& 78.17{\scriptsize$\pm$0.45}& 77.84{\scriptsize$\pm$0.62}& 73.49{\scriptsize$\pm$0.57}& 72.92{\scriptsize$\pm$0.61}& 74.58{\scriptsize$\pm$0.89}
\\
        FedProx& 80.03{\scriptsize$\pm$0.34}& 80.14{\scriptsize$\pm$0.39}& 87.29{\scriptsize$\pm$0.61}& 78.00{\scriptsize$\pm$0.52}& 77.96{\scriptsize$\pm$0.58}& 77.45{\scriptsize$\pm$0.66}& 73.28{\scriptsize$\pm$0.40}& 73.13{\scriptsize$\pm$0.69}& 74.18{\scriptsize$\pm$0.98}
\\
        FedRobust& 81.86{\scriptsize$\pm$0.39}& 81.58{\scriptsize$\pm$0.48}& 87.11{\scriptsize$\pm$0.52}& 78.24{\scriptsize$\pm$0.41}& 78.45{\scriptsize$\pm$0.70}& 80.56{\scriptsize$\pm$0.77}& 70.17{\scriptsize$\pm$0.66}& 70.38{\scriptsize$\pm$0.92}& 76.42{\scriptsize$\pm$1.22}
\\
        FedCorr& 82.10{\scriptsize$\pm$0.28}& 82.78{\scriptsize$\pm$0.34}& 85.79{\scriptsize$\pm$0.49}& 81.06{\scriptsize$\pm$0.33}& 81.31{\scriptsize$\pm$0.42}& 82.75{\scriptsize$\pm$0.56}& 74.42{\scriptsize$\pm$0.37}& 74.53{\scriptsize$\pm$0.68}& 82.04{\scriptsize$\pm$0.85}
\\
        FedGP& \underline{86.75{\scriptsize$\pm$0.19}}& \underline{85.22{\scriptsize$\pm$0.26}}& 87.53{\scriptsize$\pm$0.42}& \underline{84.98{\scriptsize$\pm$0.36}}& \underline{84.20{\scriptsize$\pm$0.51}}& 85.37{\scriptsize$\pm$0.53}& 80.15{\scriptsize$\pm$0.39}& 79.29{\scriptsize$\pm$0.52}& 81.96{\scriptsize$\pm$0.57}
\\
        FedFixer& 81.67{\scriptsize$\pm$0.34}& 82.08{\scriptsize$\pm$0.47}& 86.33{\scriptsize$\pm$0.66}& 81.12{\scriptsize$\pm$0.42}& 80.07{\scriptsize$\pm$0.53}& 85.91{\scriptsize$\pm$0.94}& 79.25{\scriptsize$\pm$0.51}& 79.90{\scriptsize$\pm$0.78}& \underline{82.95{\scriptsize$\pm$0.97}}
\\
        FedNoRo& 85.19{\scriptsize$\pm$0.31}& 84.77{\scriptsize$\pm$0.58}& \underline{88.02{\scriptsize$\pm$0.62}}& 83.89{\scriptsize$\pm$0.37}& 83.16{\scriptsize$\pm$0.65}& \underline{86.05{\scriptsize$\pm$0.45}}& \underline{80.87{\scriptsize$\pm$0.54}}& \underline{80.12{\scriptsize$\pm$0.49}}& 82.89{\scriptsize$\pm$0.81}
\\
        \pmb{Ours}& \pmb{87.54{\scriptsize$\pm$0.16}}& \pmb{87.10{\scriptsize$\pm$0.21}}& \pmb{89.12{\scriptsize$\pm$0.26}}& \pmb{85.88{\scriptsize$\pm$0.19}}& \pmb{86.75{\scriptsize$\pm$0.25}}& \pmb{88.28{\scriptsize$\pm$0.29}}& \pmb{82.08{\scriptsize$\pm$0.23}}& \pmb{81.12{\scriptsize$\pm$0.27}}& \pmb{83.28{\scriptsize$\pm$0.38}}\\
        \bottomrule
    \end{tabular}
\end{table*}
\subsubsection{Results on the Chaoyang Dataset}
\begin{figure}[htp]
\centering
\includegraphics[width=9cm]{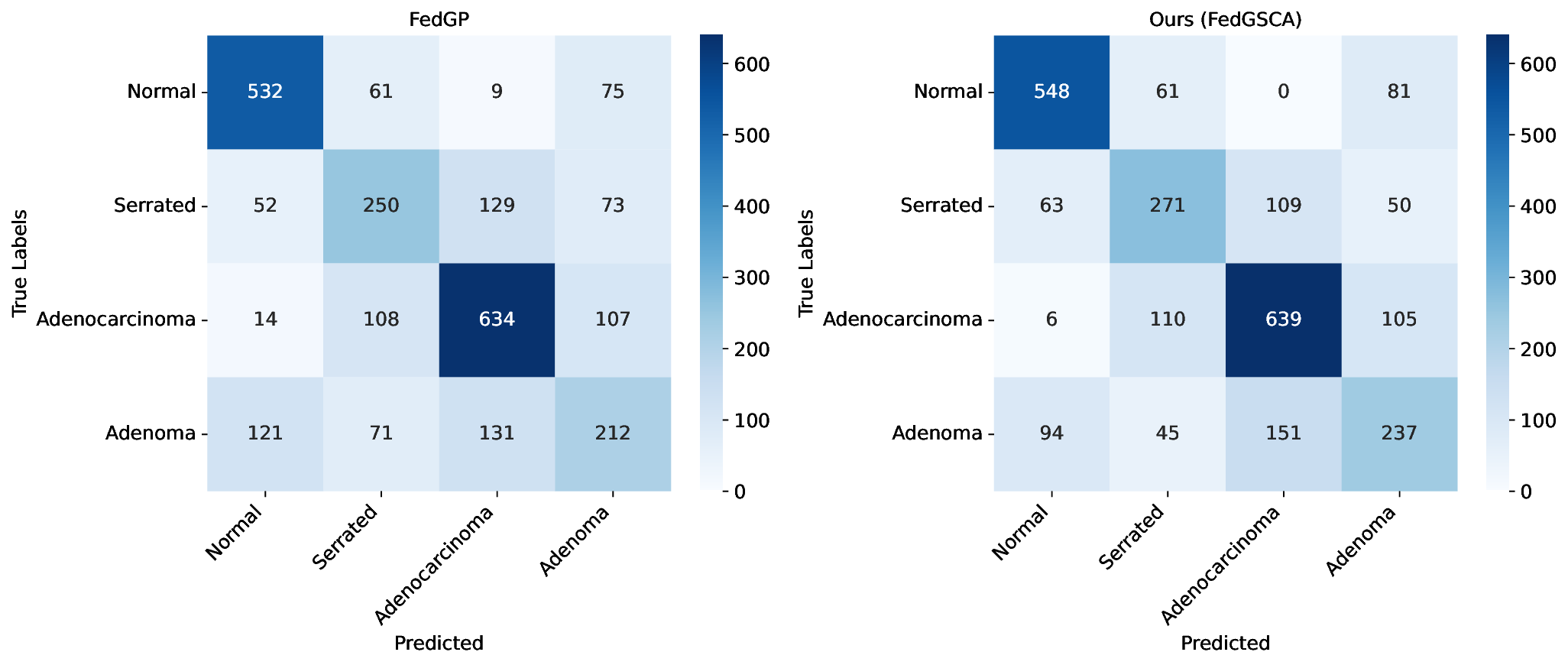}
\caption{The confusion matrices of FedGP and Ours(FedGSCA)}
\label{fig2}
\end{figure}

On the real-world Chaoyang dataset, our method achieves an F1 of 76.92, recall of 76.14, and precision of 79.05, outperforming other SOTA methods (Table \ref{table8}). As shown in Fig.\ref{fig2}, our method reduces misclassifications, particularly for the Adenocarcinoma and Adenoma classes, compared to FedGP. The results confirm that FedGSCA is effective in real-world medical federated learning scenarios.

\begin{table}[h]
    \centering
    \caption{Comparison on Chaoyang dataset. The best and second-best results are highlighted.}
    \label{table8}
    \small  % 整体缩小字号
    \setlength{\tabcolsep}{12pt}  % 调整列间距
    \renewcommand{\arraystretch}{0.8}  % 调整行高
    \begin{tabular}{@{}l*{3}{c}@{}}  % 使用紧凑列格式
        \toprule
        Method & F1 & Recall & Prec. \\
        \midrule
        FedAvg    & 69.33{\scriptsize$\pm$0.31} & 69.48{\scriptsize$\pm$0.50} & 71.14{\scriptsize$\pm$0.58} \\
        FedProx   & 71.39{\scriptsize$\pm$0.34}     & 70.22{\scriptsize$\pm$0.57}     & 72.88{\scriptsize$\pm$0.72} \\
        FedRobust & 73.81{\scriptsize$\pm$0.40}     & 73.76{\scriptsize$\pm$0.49}     & 76.37{\scriptsize$\pm$0.51} \\
        FedCorr   & 75.10{\scriptsize$\pm$0.29}     & 74.68{\scriptsize$\pm$0.35}     & 77.15{\scriptsize$\pm$0.62} \\
        FedGP     & \underline{75.98{\scriptsize$\pm$0.27}} & 75.59{\scriptsize$\pm$0.34} & 78.57{\scriptsize$\pm$0.56} \\
        FedFixer  & 74.54{\scriptsize$\pm$0.38}     & 74.91{\scriptsize$\pm$0.41}     & \underline{79.03{\scriptsize$\pm$0.49}} \\
        FedNoRo  & 75.27{\scriptsize$\pm$0.25}     & \underline{75.86{\scriptsize$\pm$0.32}}     & 78.96{\scriptsize$\pm$0.46} \\
        \pmb{Ours} & \pmb{76.92{\scriptsize$\pm$0.18}}      & \pmb{76.14{\scriptsize$\pm$0.20}}      & \pmb{79.05{\scriptsize$\pm$0.27}} \\
        \bottomrule
    \end{tabular}
\end{table}
\subsection{Ablation Analysis}
\subsubsection{Effects of tailored components}
\begin{figure}[htp]
    \centering
    \includegraphics[width=8cm]{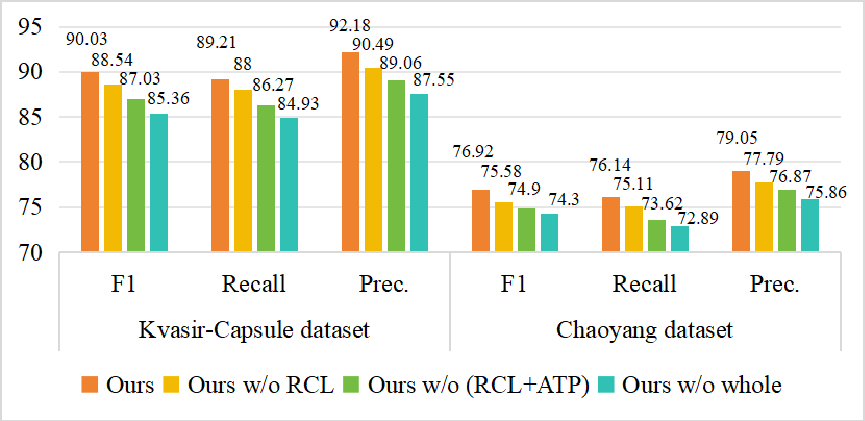}
    \caption{Ablation study on the Kvasir-Capsule dataset under 0-S10\%-P10\%-C10\% label noise configuration and the real-world Chaoyang dataset}
    \label{fig3}
\end{figure}

We conducted an ablation study on the Kvasir-Capsule and Chaoyang datasets to evaluate key components of FedGSCA, with results averaged over 5 trials. We compared our full method with three variants: Without RCL; Without RCL + ATP (fixed threshold 0.7); and Without the Whole main Mechanism, which eliminates RCL, ATP, and GSS but retains the CSS. The ablation study on the Kvasir-Capsule dataset under 0-S10\%-P10\%-C10\% label noise configuration and the real-world Chaoyang dataset is shown in Fig.\ref{fig3}. on the Kvasir-Capsule dataset. Removing the RCL component resulted in a decrease in F1 to 88.54 and precision to 90.49, showing RCL's importance, particularly in reducing false positives. Replacing the ATP with a fixed threshold led to a further decline, demonstrating that dynamic threshold is essential for accurately identifying clean samples, especially in minority classes. When the GSS component was removed, F1 falled to 85.36, recall to 84.93, and precision to 87.55. This shows that GSS plays a critical role in learning global noise knowledge, allowing the model to more accurately select clean samples. On the Chaoyang dataset, removing RCL led to an F1 drop of 1.34; removing ATP resulted in a drop of 0.68; and removing GSS resulted in a drop of 0.60. These results show the direct impact of removing these components on the model's performance.
\subsubsection{Analysis of RCL Loss}
\begin{figure}[htp]
    \centering
    \includegraphics[width=0.75\linewidth]{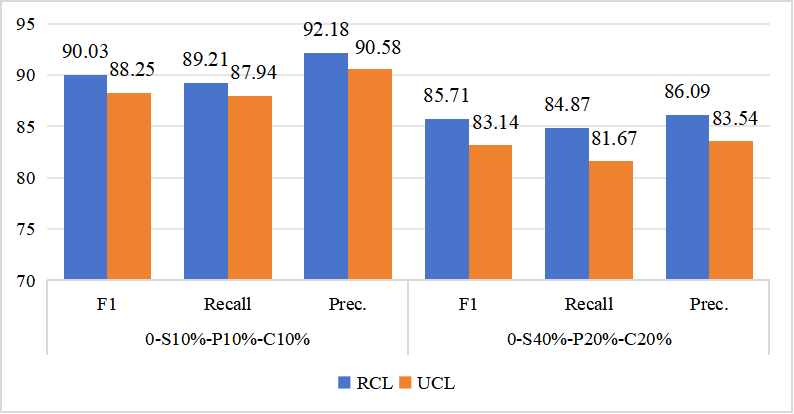}
    \caption{Comparison of RCL and UCL under varying noise levels}
    \label{fig4}
\end{figure}

We compare our method RCL to Uniform Credal Labeling (UCL). In UCL, if the model’s prediction probability $\hat{p}$ for a class different from the training label exceeds a threshold $\beta$, the label is treated as completely uncertain. Consequently, such instances are ignored in the loss optimization process, as the model does not prioritize any specific label. In contrast, RCL introduces a more refined handling of ambiguous labels by focusing on two fully plausible classes, instead of assigning equal plausibility to all possible labels. As a result, the model continues learning even from noisy or ambiguous samples, without discarding valuable information. Fig.\ref{fig4} illustrates the performance comparison, where significant improvements over UCL are observed for RCL. Incorporating credal sets with two fully plausible classes, rather than completely ignoring uncertain samples, leads to more effective loss minimization and better performance across different noise levels.
\subsubsection{Analysis of hyper-parameters}
We analyze the initial threshold $\zeta_0$ in ATP generation on the Kvasir-Capsule dataset under 0-S10\%-P10\%-C10\% noise configuration (5 trials). As shown in Table~\ref{table9:The analysis}, when $\zeta_0 = 0.8$, the model achieves optimal performance. Setting $\zeta_0$ too high (0.9) leads to fewer pseudo-labels being generated, especially for minority classes, resulting in decreased performance. Conversely, a lower threshold (0.7) increases the risk of incorrect pseudo-labels, also degrading model performance. Moreover, We further analyze the impact of the hyper-parameter $\beta$ in RCL under the 0-S10$\%$-P10$\%$-C10$\%$ label noise configuration. The results, averaged over 5 trials, are presented in Table \ref{table9:The analysis}. We set \( \beta_0 = 0.75 \) and vary \( \beta_1 \) to observe its effect on performance. The results indicate that when \( \beta_1 = 0.55 \), the model achieves the best performance. Conversely, when \( \beta_1 = 0.35 \), the performance deteriorates, suggesting that setting \( \beta_1 \) too low introduces too much uncertainty in the later stages of training, which hampers the model's ability to handle noisy labels effectively. Additionally, we compare this strategy to two alternatives: a constant $\beta$ and a linear decay, both of which show inferior results to the cosine strategy.
\begin{table}[h]
   \centering
   \caption{The analysis of hyper-parameters under 0-S10\%-P10\%-C10\% label noise configuration on OIA-ODIR dataset.}
   \label{table9:The analysis}
   \large
   \resizebox{\linewidth}{!}{
   \begin{tabular}{lcccc}
       \toprule
       \multicolumn{2}{c}{Setting}& F1& Recall&Prec.\\
       \midrule
       \multicolumn{2}{c}{$\zeta_0$ = 0.7}& 87.50& 87.02&89.05\\
       \multicolumn{2}{c}{$\zeta_0$ = 0.8}& \pmb{87.54}& \pmb{87.10}&\pmb{89.12}\\
       \multicolumn{2}{c}{$\zeta_0$ = 0.9}& 87.49& 86.93&89.01\\
       \midrule
       \multicolumn{2}{c}{Constant ($\beta$ = 0.75)}& 86.85& 86.63&88.71\\
       \multicolumn{2}{c}{Linear ($\beta_0$ = 0.75, $\beta_1$ = 0.55)}& 87.46& 87.04&89.08\\
       Cosine($\beta_0$ = 0.75)& $\beta_1$ = 0.65& 87.50& 87.04&88.99\\
       & $\beta_1$ = 0.55& \pmb{87.54}& \pmb{87.10}&\pmb{89.12}\\
       & $\beta_1$ = 0.45& 87.35& 86.87&88.53\\
       & $\beta_1$ = 0.35& 86.24& 86.45&88.51\\
       \bottomrule
   \end{tabular}}
\end{table}

\section{Discussion}
Different types of label noise in medical image annotation reflect various error patterns encountered in clinical practice. Our experiments reveal that although symmetric noise affects model performance, where each class has an equal probability of being mislabeled as any other class, it is relatively easier for the model to handle. Pairflip noise simulates a more challenging mislabeling scenario in which mislabeling occurs primarily between adjacent classes. The most challenging type is the CK-Asymm. noise, which simulates systematic error patterns in actual diagnosis. This type of noise leads to significant shifts in class distribution, particularly between frequently misdiagnosed categories. FedGSCA maintains high accuracy even under this noise setting that most closely resembles clinical practice, demonstrating its potential for real-world medical applications.

\begin{figure}[htp]
    \centering
    \includegraphics[width=0.75\linewidth]{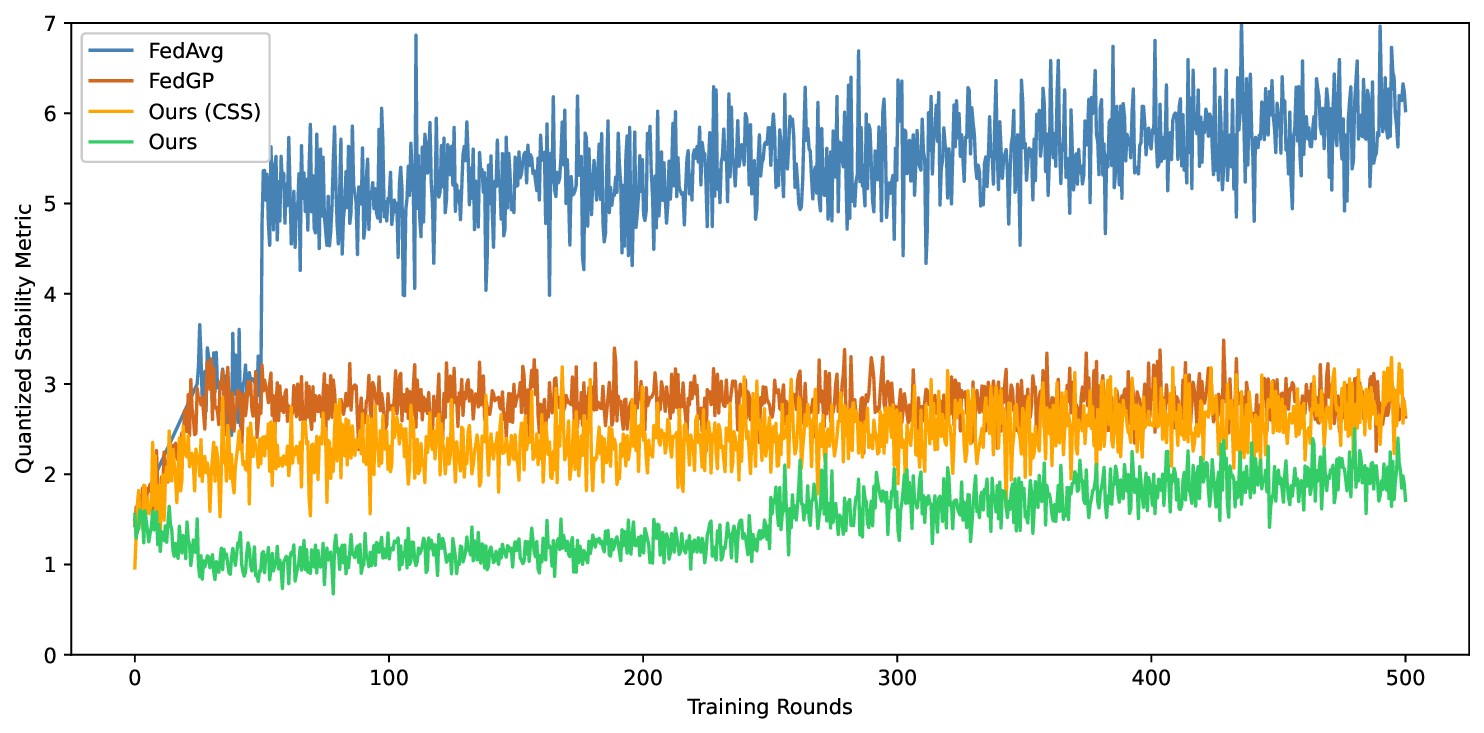}
    \caption{The evolution of quantized training stability for different methods}
    \label{fig5}
\end{figure}

To better understand the underlying challenges of our approach, we utilize Quantized Training Stability to quantify the effects of noise heterogeneity on the stability experienced during local training sessions. Specifically, we measure the stability by calculating the average proximal regularization metric between local model weights ($\theta_k^{(t)}$) and global model weights ($\theta^{(t)}$) at each communication round t:
\begin{equation}
    \text{Stability Metric} = \frac{1}{|S|} \sum_{k \in S} \|\theta_k^{(t)} - \theta^{(t)}\|^2
\end{equation}
The experiments are conducted on the Kvasir-Capsule dataset with 0-S10$\%$-P10$\%$-C10$\%$ label noise configuration. As seen in Fig.\ref{fig5}, FedAvg shows the highest instability, with the quantized stability metric increasing throughout training due to its lack of noise-handling mechanisms. In contrast, FedGP and Ours (CSS) exhibit more stable trends with smaller fluctuations. Ours (CSS)—which only employs the CSS without the GSS—slightly outperforms FedGP due to its refined noise filtering at the client level. Ours, FedGSCA, demonstrates the best stability, maintaining lower metrics across most training rounds. This reduction in instability is crucial for enhancing the performance of the federated model, particularly in scenarios with high levels of noise and data heterogeneity.

In terms of communication efficiency, we further analyze the training process on the Kvasir-Capsule dataset under the 0-S40\%-P20\%-C20\% label noise configuration. Although FedGSCA requires more time per training round (186 seconds) compared to FedAvg (133 seconds), it reaches FedAvg's peak performance level (F1 score: 77.78) in just 96.78 minutes, significantly faster than the 178.34 minutes required by FedAvg. This substantial time saving can be attributed to our GSS and CAA mechanisms, and reduces the number of training rounds needed to achieve comparable performance.
\section{Conclusion}
In this paper, we introduce FedGSCA, a novel federated learning framework designed to address the challenges of label noise and class imbalance in medical image classification. Our approach integrates the GSS to facilitate noise-handling knowledge sharing across clients, improving model robustness in the presence of noise heterogeneity. Additionally, the CAA mechanism, which includes ATP generation and RCL Loss, dynamically adjusts training strategies to fully utilize minority class samples and manage noisy labels in a cautious and flexible manner.

Comprehensive experiments on three diverse medical imaging datasets demonstrate that FedGSCA outperforms SOTA methods, confirming its robustness in extreme, heterogeneous, and real-world noise settings. Our ablation experiments further validate the importance of each component in our method. These results demonstrate that FedGSCA not only enhances model stability during local training but also significantly improves global model performance across diverse noise conditions, making it a promising solution for federated learning in real-world healthcare environments.

Future research directions may include testing FedGSCA on datasets with even more extreme heterogeneity to further validate its robustness and adaptability in highly diverse federated learning environments. Additionally, optimizing the computational efficiency of the CAA mechanism will allow FedGSCA to be more scalable and accessible in resource-constrained clinical settings. These enhancements will open new opportunities for applying federated learning frameworks in more complex medical scenarios.

\end{document}